\documentclass[11pt]{article}
\usepackage[margin=1in]{geometry}
\usepackage[utf8]{inputenc}
\usepackage{amsmath,amssymb,graphicx,booktabs,array,longtable,xcolor,setspace,enumitem}
\usepackage[authoryear,round]{natbib}
\usepackage{tikz}
\usetikzlibrary{arrows.meta}
\usepackage[colorlinks=true,linkcolor=blue!50!black,citecolor=blue!50!black,urlcolor=blue!50!black]{hyperref}
\newsavebox{\tabbox}
\newcommand{\fittab}[1]{\sbox{\tabbox}{#1}\ifdim\wd\tabbox>\textwidth\resizebox{\textwidth}{!}{\usebox{\tabbox}}\else\usebox{\tabbox}\fi}

\title{Can Respondent-Level AI Measures Recover Individual and Group-Level Associations?}
\author{Wenxin Jiang\thanks{Northwestern University (wjiang@northwestern.edu)}
\and Xuyang Wang\thanks{Nanjing University (141200073@smail.nju.edu.cn), corresponding author}
\and Yuxiao Wu\thanks{Nanjing University (yxwu2013@nju.edu.cn)}}
\date{November 2026 (version 2; supersedes version 1 of September 2026)}

\begin{document}
\maketitle

\begin{center}
\fbox{\parbox{0.93\textwidth}{\small\textbf{Notice of correction (version 2).} Version 1 of this paper (which called the approach AICOME) generated the respondent-level AI scores for the four validation dimensions in a single request per respondent that returned all concepts at once. Variable exclusions were applied only within each concept's own profile block, so a variable excluded for one concept (for example management duty, weekly hours, computer use or foreign-language use) remained visible to the model in the profile blocks of other concepts in the same request. This may have inflated the agreement between AI-derived and survey measures reported in version 1 and the conclusions built on it. In this version each dimension is scored in a separate request with its own feature set, the target variable and the variables excluded with it (Table~\ref{tab:t10}) are never in the request, scores are generated by an open-weight model (Qwen3.6-35B-A3B) instead of GPT-4o-mini, and all tables are recomputed. The abstract, the empirical sections and the conclusions have been rewritten accordingly.}}
\end{center}

\begin{abstract}
Researchers increasingly use artificial intelligence to construct measures of characteristics that conventional surveys do not record. We ask whether respondent-level AI measures can recover the individual- and group-level associations that a survey item would yield in a contextual model. In the 2022 China Family Panel Studies, with occupations as groups and four job-related survey items as benchmarks, AI measures from two prompt styles agree well with the survey items at the occupation level and partly within occupations. Performance depends on the information supplied to the model and deteriorates sharply with only occupation and basic demographics. We compare the coefficients of AI measures with those of the survey items in the standard contextual model of job satisfaction. Across 48 comparisons, no AI coefficient was significant with a sign opposite to the survey coefficient; the AI coefficients were mostly smaller, and of the comparisons with a significant survey coefficient they detected 12 of 14 at the group level, 6 of 10 at the individual level and none of 6 contextual ones. For the four validated dimensions, the associations found under both prompt styles, namely positive group associations of computer use, foreign-language use and management, and negative group and individual associations of weekly hours, are all found with the survey items as well. Applied to autonomy, people--things, creative--routine and technology, the AI measures show positive group associations for people--things and autonomy, and a positive individual association for autonomy, under both prompt styles. Such measures can screen for associations in rich existing data; they do not replace direct survey measurement. The validation framework applies to other surveys and can monitor progress as AI models become more capable.
\end{abstract}

\section{Introduction}
Many social science questions involve variables that vary both between groups and within groups. Workers are nested within occupations, students within schools, employees within firms, patients within hospitals, and residents within neighborhoods. Researchers frequently want to know whether an association reflects differences among groups, differences among individuals within the same group, or both. This distinction is central to contextual analysis. If a characteristic is associated with an outcome, does the association arise because groups differ in their average characteristics, because individuals differ from others in the same group, or because both levels matter?

AI-based measurement offers new opportunities for studying constructs that are absent from conventional surveys. Large language models can score occupations, organizations, tasks, texts, or other entities on abstract dimensions such as autonomy, technological intensity, interpersonal orientation, or creativity. Many such measures are defined at the group or entity level. For example, an AI system may assign a single autonomy score to an occupation, firm, school, or neighborhood. Such scores can be useful for studying between-group differences, but they cannot determine whether individuals within the same group differ from one another or whether such within-group deviations matter for outcomes.

At the same time, recent work on AI-assisted survey prediction, LLM-based item imputation, and virtual survey respondents shows that language models can predict missing or unasked survey responses from respondent profiles, survey-question text, temporal information, or partial respondent attributes \citep{kim2023ai,ji2024predicting,zhao2025large}. Our purpose is not to claim that respondent-level AI prediction is itself new. Instead, we ask whether AI-derived respondent-level measures can be used for a different inferential task: contextual measurement. The key question is whether an AI-derived variable can recover not only an observed survey measure, but also the individual-level and group-level associations that would be estimated if the survey measure were observed.

This paper evaluates that use and proposes a validation protocol for it. Let $g(i)$ denote the group to which individual $i$ belongs. A respondent-level AI measure $Z_i$ can be decomposed into a group mean and an individual deviation from that group mean:
\begin{equation}
Z_i=\bar Z_{g(i)}+\bigl(Z_i-\bar Z_{g(i)}\bigr).
\end{equation}
This decomposition turns a respondent-level AI measure into a tool for contextual measurement. It allows researchers to use both $Z_i$ and $\bar Z_{g(i)}$, or equivalently $\bar Z_{g(i)}$ and $Z_i-\bar Z_{g(i)}$, in models that distinguish individual-level and group-level associations. Standard group-level AI scores can measure differences across groups, but they cannot identify within-group heterogeneity because their within-group deviation is mechanically zero.

The framework is easiest to describe using an underlying-concept notation. Let $k$ index an underlying concept of theoretical interest, such as computer use, technology use, management responsibility, or foreign-language use. Let $U_{ki}$ be the true level of concept $k$ for individual $i$. Let $W_{ki}=W(U_{ki};q_k)$ denote respondent $i$'s survey-based indicator of concept $k$, where $q_k$ represents the survey-question wording, response scale, and measurement protocol. Let $Z_{ki}=Z(U_{ki},F_{ki};p_k)$ denote respondent $i$'s AI-derived indicator of the same concept, where $F_{ki}$ is the respondent-specific feature vector supplied to the AI system and $p_k$ is the prompt protocol for concept $k$. The outcome is denoted by $Y$, and regression controls are denoted by $X$. This notation emphasizes that neither $W$ nor $Z$ is treated as the true concept. They are alternative indicators of the same underlying concept, potentially shaped by different measurement protocols, response scales, and respondent-specific feature information. The AI measure is therefore not designed merely to reproduce the observed survey response. For example, a survey item may be binary while the AI prompt elicits a richer ordered score. We ask whether the contextual structure obtained from $(Z_i,\bar Z_{g(i)})$ supports substantive conclusions similar to those obtained from $(W_i,\bar W_{g(i)})$ when validation data are available.

We illustrate the approach using job satisfaction and occupational groups in the 2022 China Family Panel Studies (CFPS). In the empirical application, $Y$ is job satisfaction and $g(i)$ is occupation. The setting is useful for two reasons. First, CFPS contains rich respondent and job information that can serve as features $F$ for AI measurement. Second, CFPS contains several observed work-related variables that allow validation against conventional survey measures $W$. Occupations provide one illustration of the general grouping structure. The same logic is potentially applicable to other hierarchical settings, such as students within schools, employees within firms, patients within hospitals, and residents within neighborhoods.

The paper has three main contributions. First, we develop a validation protocol for respondent-level AI measures that goes beyond response-level correlation. Existing validation studies focus primarily on response recovery, item-level accuracy, or distributional similarity. We treat these as useful but insufficient evidence for contextual applications. Our primary validation target is recovery of contextual inference: whether AI-derived measures recover explanatory power, preserve the direction and approximate size of coefficients, reproduce individual-level and group-level conclusions, and generate similar fitted values. Coefficient-vector correlations and RMSE are reported as descriptive diagnostics, not as strict requirements that the AI measure numerically mimic the survey measure.

Second, we show that respondent-level AI measures recover the occupation-level component of contextual decompositions well and the individual component partially. Weekly hours provides the clearest case: the survey-style AI measure explains more than any occupation-level score and reproduces the sign of the within-occupation association, while occupation-level AI scores contribute almost nothing. For computer use, foreign-language use and management responsibilities the survey benchmarks themselves carry less individual-level signal than weekly hours, so recovery can be judged less sharply. Across all comparisons, AI coefficients were mostly attenuated relative to the survey coefficients but never significant with the opposite sign.

Third, we identify boundary conditions. Performance depends on the information supplied: with only occupation and basic demographics, the average within-occupation correlation falls from about 0.26 to about 0.10. Recovery is also partial when several related concepts are simultaneously unobserved. The later application to autonomy, people--things, creative--routine, and technology illustrates the intended use, but because those latent dimensions lack direct CFPS validation measures, those results are interpreted as an application rather than as direct proof of latent truth.

\begin{figure}[t]
\centering
\begin{tikzpicture}[
 box/.style={draw,rounded corners=2pt,align=center,font=\footnotesize,inner sep=4pt,text width=4.4cm,minimum height=9mm},
 wbox/.style={draw,rounded corners=2pt,align=center,font=\footnotesize,inner sep=4pt,text width=11.2cm,minimum height=9mm},
 arr/.style={-{Latex[length=2mm]},thick}]
\node[box,fill=gray!12] (lit) at (-5.0,0) {\textbf{Existing literature}\\AI-augmented surveys, item-nonresponse imputation, virtual respondents};
\node[box,fill=gray!12] (tgt) at (-5.0,-2.0) {\textbf{Typical validation target}\\response recovery};
\draw[arr] (lit)--(tgt);
\node[box,fill=blue!6,text width=6.4cm] (U) at (2.3,0) {\textbf{Underlying concept} $U_{ki}$\\computer use, weekly hours, management, foreign-language use};
\node[box,fill=white,text width=3.0cm] (W) at (0.2,-2.0) {\textbf{Survey indicator}\\$W_{ki}=W(U_{ki};q_k)$};
\node[box,fill=white,text width=4.6cm] (Z) at (4.6,-2.0) {\textbf{AI indicator} $Z_{ki}=Z(U_{ki},F_{ki};p_k)$\\one dimension per request; excluded variables absent};
\draw[arr] (U)--(W); \draw[arr] (U)--(Z);
\node[wbox,fill=green!8] (dec) at (2.0,-4.0) {\textbf{Decomposition of each indicator} \mbox{$M\in\{W,Z\}$}\\[1pt] individual value \mbox{$M_{ki}$} $\cdot$ group mean \mbox{$\bar M_{k,g(i)}$} $\cdot$ individual deviation \mbox{$M_{ki}-\bar M_{k,g(i)}$}};
\draw[arr] (W)--(W|-dec.north); \draw[arr] (Z)--(Z|-dec.north);
\node[wbox,fill=yellow!10] (cm) at (2.0,-5.75) {\textbf{Contextual model}, fitted with \mbox{$M=W$} and with \mbox{$M=Z$}:\\[1pt] $\beta_I$ (individual), $\beta_C$ (contextual), \mbox{$\beta_G=\beta_C+\beta_I$} (group)};
\draw[arr] (dec)--(cm);
\node[wbox,fill=orange!8] (val) at (2.0,-7.45) {\textbf{Validation}: compare the coefficients of $Z$ with those of $W$\\ (sign-detection, sign-conflict, sign-surplus, size ratio; Table~\ref{tab:t6a})\\ response-level agreement $\cdot$ model-level $R^2$ $\cdot$ boundary conditions};
\draw[arr] (cm)--(val);
\node[wbox,fill=red!6] (goal) at (2.0,-9.0) {\textbf{Aim}: evaluate how much contextual inference is preserved, not merely whether responses are reproduced};
\draw[arr] (val)--(goal);
\end{tikzpicture}
\caption{Conceptual framework for evaluating respondent-level AI measures in contextual analysis. Survey variables and AI-derived variables are treated as alternative indicators of an underlying concept. Each indicator is decomposed into a group mean and an individual deviation, and the contextual model is fitted with each, giving group, contextual and individual associations. Validation compares the coefficients obtained with the AI measure with those obtained with the survey item.}
\label{fig:framework}
\end{figure}

Figure~\ref{fig:framework} summarizes the central distinction between the present study and recent AI-assisted survey-prediction research. Existing work has primarily evaluated the ability of large language models to recover missing responses, respondent attributes, or population distributions. Our focus instead is whether respondent-level AI measures can support contextual analysis. Consequently, the principal validation target is not response recovery alone, but recovery of between- and within-occupation inferences.

\section{Related Literature and Positioning}
\subsection{Contextual models and occupational measurement}
A central concern in the contextual-model literature is distinguishing between-group and within-group associations \citep{blalock1984,enders2007}. Researchers studying neighborhoods, schools, firms, and occupations often seek to determine whether outcomes are driven by group environments or by differences among individuals within the same group. Estimating such models requires measures that vary at the individual level. Group-level measures can be useful for describing between-group differences, but by construction they cannot identify within-group deviations.

A large occupational-measurement literature has developed resources such as the Dictionary of Occupational Titles and O*NET to characterize jobs and occupations \citep{peterson2001,autor2003,acemoglu2011,deming2017}. These resources have been widely used to measure task content, skill requirements, technological exposure, social skills, and other features of work. Their central strength is that they provide systematic and comparable measures across occupations. Their central limitation for the present purpose is that they are usually occupation-level measures. If all workers in the same occupation receive the same score, the measure can identify only between-occupation variation.

Recent advances in large language models have created a related AI-measurement literature. Researchers have used language models to score occupations, estimate exposure to artificial intelligence, classify work tasks, generate synthetic respondents, and construct new social-science measures \citep{felten2021,eloundou2024}. These approaches overlap with the present paper because they use AI systems to generate quantities that were not directly measured in conventional datasets. However, most existing AI occupational measures remain group-level: occupations, occupational tasks, industries, firms, or texts receive scores, and individuals inherit the score attached to their group or label. Our interest is in respondent-level measures that can be decomposed into occupation means and within-occupation deviations, making contextual analysis possible.

\subsection{AI survey prediction, imputation, and synthetic respondents}
Individual AI measurement is related to several emerging uses of large language models in simulating human behavior and in survey research. \citet{horton2023} use LLMs as simulated economic agents to reproduce and explore behavioral responses in hypothetical experiments. \citet{argyle2023} use LLMs to generate synthetic respondents (``silicon samples'') and evaluate whether the resulting responses reproduce patterns observed in human survey data. Another line of work uses LLMs to augment surveys, address item nonresponse, or act as virtual survey respondents \citep{kim2023ai,ji2024predicting,zhao2025large}.

These studies overlap with the present paper because all use AI systems to infer quantities not directly observed in a survey. The distinction is in the estimand and validation target. Prior work primarily asks whether AI systems can reproduce responses, opinions, attributes, or distributions. We ask whether respondent-level AI-derived measures can support contextual analysis. Specifically, we evaluate whether AI-generated variables recover explanatory power, preserve coefficient directions, and reproduce substantive between-within conclusions obtained from observed CFPS measures. The validation target is therefore not item-level prediction alone, but recovery of substantive contextual inferences.

This distinction is important because high response-level similarity need not imply valid downstream inference. Synthetic survey data may match marginal averages while producing different variation or regression coefficients, and results may vary with prompt wording or model changes \citep{bisbee2024}. Motivated by this concern, our validation strategy compares not only correlations between AI-derived and observed measures, but also incremental explanatory power, coefficient directions, contextual conclusions, coefficient-vector similarity, and fitted-value similarity.

\subsection{Construct validation, generated regressors, and measurement error}
The validation problem in the present study is related to but not identical to either a classical generated-regressor problem or a simple measurement-error problem. In generated-regressor settings, a first-stage estimate is often assumed to converge to a well-defined target regressor, and the central issue is how first-stage estimation uncertainty affects second-stage standard errors \citep{pagan1984,murphy1985}. See also \citet{escanciano2023} for an automatic locally robust GMM framework for inference with machine-learning-generated regressors. Recently, \citet{ludwig2026} develop an econometric framework for using LLM outputs in prediction and estimation, emphasizing that valid downstream inference for LLM-measured concepts requires validation data to debias and account for LLM errors. In the present setting, the AI-derived measure is not assumed to converge to the observed survey response as sample size grows. Nor do we assume that the observed survey response is a perfect measure of the underlying job concept that affects satisfaction. Both the observed survey variable and the AI-derived variable may be imperfect indicators of an underlying concept.

For this reason, the central validation question is whether the AI-derived measure yields substantively similar inferences to the observed survey measure when no perfect validation data are available. This framing is closer to construct validation than to a narrow two-step inference correction. The construct-validity tradition emphasizes that validity is not established by a single statistic, but by evidence that the operational measure supports the intended interpretation and use \citep{cronbach1955,messick1989,adcock2001}. Here the intended use is contextual analysis. We therefore evaluate whether AI-derived measures reproduce the contextual inferences obtained from conventional survey measures.

\emph{Target and benchmark.} Let $U$ denote the concept described by the survey item or AI prompt, such as computer use, management responsibility, weekly hours or foreign-language use. The quantity of substantive interest is the concept's group and individual association with the outcome in the specification that practitioners use: the contextual model of Section~3.2, with the standard controls entered once. Since $U$ is not observed, we take the coefficients of the survey item $W$ in the same specification as the benchmark for that target, and ask how closely the coefficients of the AI measure $Z$ match them in sign and size. We make no assumption that either indicator is a valid proxy for $U$ in a formal sense; the comparison is empirical. Formal conditions are beyond the scope of this paper.

Both indicators can fall short of the concept in different ways. A dichotomous survey item and a five-point AI score each carry at most part of a continuous concept, and the AI score also depends on the features supplied to the model. Individual coefficient signs, coefficient-vector correlations, and RMSE are therefore treated as diagnostics, not as requirements that the AI measure numerically mimic the survey measure.

\section{Measurement and Validation Framework}
\subsection{Group-level AI scores}
A common measurement strategy is to assign a score directly to a group or entity. In an occupational application, this means assigning each occupation a score for a dimension such as technology or autonomy. In other settings, the group might be a firm, school, hospital, neighborhood, or region. If individual $i$ belongs to group $g(i)$ and the AI score is constructed directly at the group level, the measure takes the form $\hat z_i=\hat z_{g(i)}$. This measure can be used to estimate how outcomes differ across groups. But because every member of the same group has the same value, the within-group deviation is zero, $\hat z_i-\hat z_{g(i)}=0$. Thus, group-level AI scores cannot test whether within-group heterogeneity exists. The respondent-level approach differs by first constructing respondent-level measures and then deriving group-level aggregates from those respondent-level measurements.

\subsection{Respondent-level AI measures}
The approach begins by constructing respondent-level indicators of an underlying concept. For $U_{ki}$, the level of concept $k$ for individual $i$, the AI-derived measure is written $Z_{ki}=Z(U_{ki},F_{ki};p_k)$, where $F_{ki}$ denotes respondent $i$'s feature vector for concept $k$ and $p_k$ denotes the prompting protocol. In the CFPS application, the feature set always includes group membership (if available), represented by occupation, but the remaining features vary across target concepts. Each dimension is scored in its own request, which contains only that dimension's allowed features: variables excluded for a dimension (and, in the simultaneous-missingness design, for all four validation dimensions) are absent from the request. The rich-prompt strategy uses a richer description of the concept, while the survey-prompt strategy uses a more concise prompt structured around the questionnaire item used in the actual CFPS survey. Both strategies yield respondent-level predicted scores for each dimension.

Because these scores vary within groups, they can be aggregated to group means and used in contextual models, paralleling standard centering logic in multilevel and contextual models \citep{enders2007}. The group mean is
\begin{equation}
\bar Z_{k,g}=\frac1{n_g}\sum_{i:g(i)=g}Z_{ki},\qquad n_g=\sum_{i:g(i)=g}1 .
\end{equation}
The contextual model can be written either in the raw contextual parameterization as
\begin{equation}
Y_i=\sum_k\beta_{Ik}Z_{ki}+\sum_k\beta_{Ck}\bar Z_{k,g(i)}+X_i'\gamma+\varepsilon_i,
\end{equation}
or in the deviation contextual parameterization as
\begin{equation}
Y_i=\sum_k\beta_{Ik}\bigl(Z_{ki}-\bar Z_{k,g(i)}\bigr)+\sum_k\beta_{Gk}\bar Z_{k,g(i)}+X_i'\gamma+\varepsilon_i,
\end{equation}
where $Y_i$ is the outcome, $Z_{ki}$ is the respondent-level AI measure for dimension $k$, $\bar Z_{k,g(i)}$ is the corresponding group mean, and $X_i$ is a vector of controls. In the CFPS application, $Y_i$ is job satisfaction and $g(i)$ is occupation. We refer to $\beta_I$ as the individual or within-group association, $\beta_C$ as the contextual contrast, and $\beta_G$ as the group or between-group coefficient. The three regression coefficients satisfy
\begin{equation}
\beta_G=\beta_I+\beta_C .
\end{equation}
When the $\beta$'s have the same sign, $\beta_C$ is smaller in magnitude and harder to detect than $\beta_G$. This can also be aggravated by variance inflation: the standard error of $\beta_C$ is often larger than that of $\beta_G$, since the group means $\bar Z_{k,g(i)}$ are likely to be more associated linearly with the other right-hand-side variables $(Z_{ki},X_i)$ in the raw parameterization than with $(Z_{ki}-\bar Z_{k,g(i)},X_i)$ in the deviation parameterization.

All reported regression coefficients are standardized: $Y_i$ and $X_i$ components are standardized by their respective standard deviations, and the raw $Z_{ki}$, its group mean $\bar Z_{k,g(i)}$ and its deviation from the group mean $Z_{ki}-\bar Z_{k,g(i)}$ are all standardized by the standard deviation of the raw $Z_{ki}$, so that $\beta_G=\beta_I+\beta_C$ holds. Standard errors are clustered by occupation, with the usual finite-sample adjustment; this treats occupations as independent and relies on a large number of them. Occupation means are computed on each model's estimation sample. Here $\beta_G$ is the coefficient of the occupation mean among the surveyed respondents, not of the latent occupation mean in the workforce; with finite occupation sizes these define different coefficients, not merely ones estimated with different sampling errors \citep[see, e.g.,][]{ludtke2008}.

The controls $X_i$ in the CFPS application include age, gender, years of education, an indicator for being married, urban residence, agricultural hukou, public or state-owned employment, whether the job is outdoors, health, and log income. Weekly hours is excluded from the control set when weekly hours is a focal validated dimension, and included otherwise.

\subsection{Validation framework}
The empirical validation proceeds at four levels. First, response-level validation examines correlations between AI-derived measures $Z$ and observed survey measures $W$. These correlations may suggest shared concept-related signal, but are not the final validation target. Second, model-level validation asks whether AI-derived measures recover incremental explanatory power and coefficient directions in non-contextual regressions. Third, contextual validation asks whether AI-derived measures recover the individual-level and group-level conclusions obtained from survey measures. Fourth, boundary validation examines when the approach deteriorates, using reduced-information and simultaneous-missingness designs.

\section{Data and Measures}
\subsection{Data}
The empirical illustration uses the 2022 wave of the China Family Panel Studies (CFPS). The data are from the China Family Panel Studies (CFPS), funded by the 985 Program of Peking University and carried out by the Institute of Social Science Survey of Peking University. CFPS is useful for this validation exercise because it contains rich respondent information, occupational membership, job-related characteristics, and subjective well-being measures. These features make it possible to construct respondent-level AI measures, aggregate them within occupational groups, and validate the resulting contextual structure against directly observed survey measures.

The outcome variable $Y_i$ is job satisfaction. In the present application, the grouping variable $g(i)$ is occupation, identified using cleaned occupational codes. Occupations are used as the empirical contextual unit for constructing group-level aggregates and estimating contextual models. The methodological framework, however, is not occupation-specific. The same logic applies to other settings in which individuals are nested within larger social units.

The validation analysis uses four observed work-related CFPS dimensions: computer use, foreign-language use, weekly hours, and management responsibilities. In the notation of the framework, these observed survey variables are denoted by $W$. They provide benchmarks for evaluating AI-derived measures $Z$. The analysis sample starts from the 27{,}001 respondents in the cleaned individual-level CFPS 2022 file and keeps those with non-missing job satisfaction, cleaned occupation code and the ten baseline controls of Section~3.2 (6{,}920 respondents). Excluding the 24 respondents with missing weekly hours, which is a control in the latent analyses (Appendix~\ref{app:A}), leaves 6{,}896 respondents in 309 occupations, for whom the AI measures were generated. The estimation samples, complete cases on the outcome, the controls and the measures of each model, contain 6{,}409 to 6{,}896 respondents in 304 to 309 occupations (Tables~\ref{tab:t3}--\ref{tab:t8}). The later latent application examines autonomy, people--things, creative--routine, and technology, which are not directly observed in CFPS and therefore cannot be validated in the same way. Detailed variable definitions, cleaning rules, AI feature sets $F$, and prompt protocols $p$ are documented in the appendices.

\subsection{AI-derived measures}
For each observed CFPS dimension, we compare the survey measure $W$, the rich-prompt AI measure $Z^{\mathrm{rich}}$, and the survey-prompt AI measure $Z^{\mathrm{survey}}$. We also compare individual AI measures with occupation-mean versions of those measures, occupation-level AI scores, and embedding-based occupation scores in non-contextual validation models. For contextual validation, occupation-level scores cannot identify within-occupation deviations, so the contextual comparisons focus on survey respondent-level measures, rich-prompt individual AI measures, and survey-prompt individual AI measures.

All AI scores in this version (individual and occupation-title) are generated by Qwen3.6-35B-A3B, an open-weight model, in an 8-bit quantization. The model ran locally with MLX-VLM 0.7.10 on a laptop computer (Apple M4 Max, 64 GB memory) that was disconnected from the internet during scoring, with greedy decoding (temperature 0) and a strict JSON output format; Appendix~\ref{app:E} gives the prompts and settings. Each dimension is scored in a separate request containing only its allowed features. The individual-level run comprised 248,256 valid requests, and the occupation-title run 1,218 valid requests, one for each occupation in the concept's sample of Table~\ref{tab:t3}. Occupation embeddings are unchanged from version 1. Requests were decoded serially, one request at a time.

For the latent application, we examine autonomy, people--things, creative--routine, and technology. These dimensions are not directly observed in CFPS. Therefore, the latent analysis evaluates robustness across rich-prompt and survey-prompt AI measures rather than comparison with survey measures. It also serves as an illustration of how respondent-level AI measures can be used for contextual analysis of a concept that is not included in an actual survey.

\section{Validation Using CFPS-Measured Dimensions}\label{sec:valid}
\subsection{Correlation validation of AI-derived measures}
Before examining contextual models, we first assess the correspondence between the AI-derived measures and the observed CFPS variables used for validation. Table~\ref{tab:t1} reports the correlations between the survey measures and the corresponding rich-prompt and survey-prompt AI measures. Since in the raw or deviation parameterization of the contextual models each variable may appear on the right-hand side in three forms (the raw $Z_{ki}$, its group mean $\bar Z_{k,g(i)}$, and the deviation from the group mean $Z_{ki}-\bar Z_{k,g(i)}$), we report the correlations with the corresponding CFPS variable in all three forms.

Occupation means are recovered well: the correlations between AI and survey occupation means range from 0.67 to 0.90 across dimensions and prompt styles, highest for computer use and management (Table~\ref{tab:t1}). Individual-level correlations are lower, and correlations of within-occupation deviations are lower still, from 0.16 to 0.36, highest for weekly hours. Management, whose within-occupation variation is hardest to infer from the available features, has the lowest deviation correlations. These correlations indicate shared signal between the AI and survey indicators, chiefly at the occupation level, and are not the final validation target.

\begin{table}[htbp]\centering\footnotesize\renewcommand{\thetable}{1}
\caption{Validation correlations between AI-derived and survey measures}\label{tab:t1}
\par\medskip\noindent\textit{Correlations}\par\smallskip
\fittab{\begin{tabular}{lcccccc}\toprule
Construct & Rich $Z$ vs.\ $W$ & Rich $\bar Z$ vs.\ $\bar W$ & Rich deviations & Survey $Z$ vs.\ $W$ & Survey $\bar Z$ vs.\ $\bar W$ & Survey deviations \\ \midrule
Computer use & 0.6284 & 0.8968 & 0.3001 & 0.6080 & 0.8804 & 0.2660 \\
Foreign language & 0.3457 & 0.6830 & 0.2300 & 0.3797 & 0.7888 & 0.2282 \\
Weekly hours & 0.4161 & 0.6702 & 0.3480 & 0.4482 & 0.7722 & 0.3585 \\
Management & 0.4169 & 0.8767 & 0.1593 & 0.4157 & 0.8660 & 0.1698 \\
\bottomrule\end{tabular}}
\par\medskip\noindent\textit{Pairwise complete-case N}\par\smallskip
\fittab{\begin{tabular}{lcccccc}\toprule
Construct & Rich $Z$ vs.\ $W$ & Rich $\bar Z$ vs.\ $\bar W$ & Rich deviations & Survey $Z$ vs.\ $W$ & Survey $\bar Z$ vs.\ $\bar W$ & Survey deviations \\ \midrule
Computer use & 6521 & 6894 & 6521 & 6521 & 6894 & 6521 \\
Foreign language & 6521 & 6894 & 6521 & 6521 & 6894 & 6521 \\
Weekly hours & 6896 & 6896 & 6896 & 6896 & 6896 & 6896 \\
Management & 6517 & 6894 & 6517 & 6517 & 6894 & 6517 \\
\bottomrule\end{tabular}}
\par\smallskip\begin{minipage}{0.95\textwidth}\footnotesize Correlations are Pearson; occupation-mean correlations are respondent-weighted; $N$ is the pairwise complete-case sample.\end{minipage}
\end{table}

\subsection{Sensitivity to available respondent information}
The correlations in Table~\ref{tab:t1} naturally raise a question about the information available to the AI model. The approach does not generate information from nothing: it constructs indicators of an underlying concept from respondent characteristics supplied in the prompt. Consequently, its usefulness depends on the amount of concept-relevant information contained in the available covariates. To evaluate this dependence, we compare two information sets. The first is the full-information specification used throughout the main analysis. The second is a reduced-information specification intended to mimic a shorter survey containing only occupation and basic demographic characteristics:
\[
F_{\mathrm{reduced}}=\{\text{occupation, age, gender, education degree, marital status, urban residence}\}.
\]
This specification excludes all job-specific information. The full information set varies by target dimension and is intentionally designed to use information that would typically be available in a labor-force or household survey; the feature sets are listed in Table~\ref{tab:t10} (Appendix~\ref{app:B}). Importantly, these full-information specifications are not constructed from variables that are inherently unavailable in survey settings. Most of the variables, such as occupation, industry, education, age, gender, income, job tenure, employer type, supervisory responsibilities, organizational size, and working hours, are routinely collected in labor-force and household surveys, while more detailed employment characteristics, such as work schedules and promotion history, are available in more detailed surveys. The full-information specification therefore represents a rich but empirically realistic survey environment in which one survey item is missing and must be inferred from other observed respondent characteristics.

Restricting the information lowers all correlations (Table~\ref{tab:t2}). Averaged over the four dimensions, the within-occupation deviation correlation falls from about 0.26 to about 0.10 under either prompt (rows ``Average''), while the individual and occupation-mean correlations fall less. The decline is sharpest for weekly hours, where even the occupation-mean correlation drops markedly. Computer use and management keep moderate individual and high occupation-mean correlations, because occupation carries much of their signal, while their within-occupation deviation correlations also fall. The approach therefore depends on informative respondent characteristics, and the within-occupation component is the part most sensitive to the loss of job-specific information. When richer job-related information is available, the AI-derived measures track the observed survey variables more closely; when it is sparse, performance deteriorates sharply. The method should not be interpreted as creating respondent-level information that is absent from the observed data. This boundary condition is consistent with recent work on augmenting survey data with generative AI: \citet{brynjolfsson2026} find that supplying LLMs with rich contextual information beyond demographics substantially improves predictive accuracy in economic survey applications, whereas changes to prompting strategy alone yield smaller gains.

\begin{table}[htbp]\centering\footnotesize\renewcommand{\thetable}{2}
\caption{Sensitivity of validation correlations to respondent information (full versus reduced information)}\label{tab:t2}
\par\medskip\noindent\textit{Rich --- Correlations}\par\smallskip
\fittab{\begin{tabular}{lcccccc}\toprule
Construct & Z Full & Z Reduced & $\bar Z$ Full & $\bar Z$ Reduced & deviation Full & deviation Reduced \\ \midrule
Computer use & 0.6284 & 0.5756 & 0.8968 & 0.8565 & 0.3001 & 0.1796 \\
Foreign language & 0.3457 & 0.2286 & 0.6830 & 0.5876 & 0.2300 & 0.0539 \\
Weekly hours & 0.4161 & 0.0992 & 0.6702 & 0.1388 & 0.3480 & 0.1137 \\
Management & 0.4169 & 0.3856 & 0.8767 & 0.8593 & 0.1593 & 0.0732 \\
Average & 0.4518 & 0.3223 & 0.7817 & 0.6106 & 0.2593 & 0.1051 \\
\bottomrule\end{tabular}}
\par\medskip\noindent\textit{Survey --- Correlations}\par\smallskip
\fittab{\begin{tabular}{lcccccc}\toprule
Construct & Z Full & Z Reduced & $\bar Z$ Full & $\bar Z$ Reduced & deviation Full & deviation Reduced \\ \midrule
Computer use & 0.6080 & 0.5595 & 0.8804 & 0.8355 & 0.2660 & 0.1842 \\
Foreign language & 0.3797 & 0.3046 & 0.7888 & 0.7588 & 0.2282 & 0.0766 \\
Weekly hours & 0.4482 & 0.1478 & 0.7722 & 0.3219 & 0.3585 & 0.0799 \\
Management & 0.4157 & 0.3869 & 0.8660 & 0.8549 & 0.1698 & 0.0754 \\
Average & 0.4629 & 0.3497 & 0.8269 & 0.6928 & 0.2556 & 0.1040 \\
\bottomrule\end{tabular}}
\par\medskip\noindent\textit{Rich --- Pairwise complete-case N}\par\smallskip
\fittab{\begin{tabular}{lcccccc}\toprule
Construct & Z Full & Z Reduced & $\bar Z$ Full & $\bar Z$ Reduced & deviation Full & deviation Reduced \\ \midrule
Computer use & 6521 & 6521 & 6894 & 6894 & 6521 & 6521 \\
Foreign language & 6521 & 6521 & 6894 & 6894 & 6521 & 6521 \\
Weekly hours & 6896 & 6896 & 6896 & 6896 & 6896 & 6896 \\
Management & 6517 & 6517 & 6894 & 6894 & 6517 & 6517 \\
\bottomrule\end{tabular}}
\par\medskip\noindent\textit{Survey --- Pairwise complete-case N}\par\smallskip
\fittab{\begin{tabular}{lcccccc}\toprule
Construct & Z Full & Z Reduced & $\bar Z$ Full & $\bar Z$ Reduced & deviation Full & deviation Reduced \\ \midrule
Computer use & 6521 & 6521 & 6894 & 6894 & 6521 & 6521 \\
Foreign language & 6521 & 6521 & 6894 & 6894 & 6521 & 6521 \\
Weekly hours & 6896 & 6896 & 6896 & 6896 & 6896 & 6896 \\
Management & 6517 & 6517 & 6894 & 6894 & 6517 & 6517 \\
\bottomrule\end{tabular}}
\par\smallskip\begin{minipage}{0.95\textwidth}\footnotesize Full: all allowed features for the dimension. Reduced: occupation, age, gender, education degree, marital status and urban residence only. Averages are signed arithmetic means of the four correlations.\end{minipage}
\end{table}

\subsection{Single-dimension non-contextual validation}
Correlation alone is not the primary criterion for evaluating the usefulness of the AI-derived measures. The remainder of the validation section focuses on a more demanding test: whether AI-derived measures reproduce explanatory power, contextual conclusions, and multivariate regression results obtained from the observed CFPS variables.

We begin with non-contextual models that add one measured job dimension at a time (Table~\ref{tab:t3}). This first exercise compares several measurement strategies: individual AI measures, occupation means of the survey and AI measures, occupation-level AI scores, and embedding-based scores (the two occupation-level benchmarks are described in Appendix~\ref{app:F}). We emphasize the gain in $R^2$ relative to the controls-only model because it measures how much each indicator adds to the explanation of the outcome beyond the controls.

The clearest case is weekly hours (Table~\ref{tab:t3}). The individual AI measures raise $R^2$ less than the CFPS item, the survey-prompt measure more than the rich-prompt one, and the survey-prompt individual measure explains more than any occupation-level measure, including the CFPS occupation mean. This illustrates why individual-level measurement can matter when a construct has substantial within-occupation signal. For computer use, foreign-language use and management every gain in $R^2$ is small, and we caution against ranking the measures in those dimensions because of sampling variation; in computer use and foreign language the survey-prompt AI measure even exceeds the gain of the CFPS item.

\begin{table}[htbp]\centering\footnotesize\renewcommand{\thetable}{3}
\caption{Single-dimension non-contextual validation: gain in $R^2$ from alternative measures}\label{tab:t3}
\fittab{\begin{tabular}{lcccc}\toprule
Gain in $R^2$ over the controls-only model & Computer use & Foreign language & Weekly hours & Management \\ \midrule
\multicolumn{5}{l}{\textit{CFPS item}} \\
\quad Individual measure & 0.0028 & 0.0004 & 0.0193 & 0.0042 \\
\quad Occupation mean & 0.0030 & 0.0008 & 0.0059 & 0.0022 \\
\addlinespace[3pt]
\multicolumn{5}{l}{\textit{Rich-prompt AI}} \\
\quad Individual measure & 0.0022 & 0.0006 & 0.0040 & 0.0017 \\
\quad Occupation mean & 0.0017 & 0.0008 & 0.0013 & 0.0017 \\
\addlinespace[3pt]
\multicolumn{5}{l}{\textit{Survey-prompt AI}} \\
\quad Individual measure & 0.0037 & 0.0013 & 0.0091 & 0.0029 \\
\quad Occupation mean & 0.0034 & 0.0009 & 0.0041 & 0.0023 \\
\addlinespace[3pt]
\multicolumn{5}{l}{\textit{Occupation-title measures}} \\
\quad AI score & 0.0027 & 0.0003 & 0.0001 & 0.0020 \\
\quad Embedding score & 0.0011 & 0.0003 & 0.0018 & 0.0028 \\
\midrule
Controls-only $R^2$ & 0.0759 & 0.0759 & 0.0541 & 0.0756 \\
$N$ & 6413 & 6413 & 6774 & 6409 \\
Occupations $J$ & 304 & 304 & 306 & 304 \\
\bottomrule\end{tabular}}
\par\smallskip\begin{minipage}{0.95\textwidth}\footnotesize Each entry is the increase in $R^2$ when the measure is added to the controls-only model, computed from unrounded values; the $R^2$ of each model is the controls-only $R^2$ plus the gain. Small differences between measures should not be used to rank them, because of sampling variation.\end{minipage}
\end{table}

\subsection{Single-dimension contextual validation}
The next validation step asks whether AI-derived individual measures recover the contextual decomposition obtained from observed CFPS measures. For each construct, we estimate the raw contextual parameterization
\begin{equation}
Y_i=\beta_Iz_i+\beta_C\bar z_{\mathrm{occ}(i)}+X_i'\gamma+\varepsilon_i,
\end{equation}
or the equivalent deviation parameterization
\begin{equation}
Y_i=\beta_I\bigl(z_i-\bar z_{\mathrm{occ}(i)}\bigr)+\beta_G\bar z_{\mathrm{occ}(i)}+X_i'\gamma+\varepsilon_i,
\end{equation}
where $z_i$ is the individual measure and $\bar z_{\mathrm{occ}(i)}$ is the corresponding occupation mean. In these parameterizations, $\beta_I$ is the individual or within-occupation coefficient, $\beta_C$ is the contextual contrast associated with the occupation mean, and the group or between-occupation coefficient is $\beta_G=\beta_I+\beta_C$. Table~\ref{tab:t4} reports the results.

Table~\ref{tab:t4} marks the coefficients that are significant at the 5\% level. The AI measures reproduce the signs of the group and individual associations in all four dimensions. Between occupations, both AI measures give significant coefficients with the CFPS sign in all four dimensions, smaller than the CFPS ones. Within occupations, both AI measures detect the negative association of weekly hours, with coefficients smaller than the CFPS one, the survey-prompt coefficient closer to it; the survey-prompt AI also detects the positive within-occupation association of management, and a positive one for foreign language that is not significant for CFPS. For computer use the individual AI coefficients are larger than the CFPS one but not significant. CFPS has significant contextual coefficients for computer use, weekly hours and management; no AI contextual coefficient is significant. The survey-prompt AI yields a larger incremental $R^2$ than the rich-prompt AI in all four dimensions, close to or above that of CFPS for computer use and foreign language and well below it for weekly hours. Because proxies can be attenuated by different amounts at the individual and group levels, we interpret sign concordance, incremental $R^2$ and substantive conclusions as the primary evidence rather than requiring exact coefficient equality.

\begin{table}[htbp]\centering\footnotesize\renewcommand{\thetable}{4}
\caption{Single-dimension contextual validation using CFPS-measured dimensions}\label{tab:t4}
\par\medskip\noindent\textit{Estimates}\par\smallskip
\fittab{\begin{tabular}{lccccc}\toprule
Construct & Method & $R^2$ & $\beta_G$ (SE) & $\beta_C$ (SE) & $\beta_I$ (SE) \\ \midrule
Computer use & CFPS & 0.0799 & 0.126 (0.029)$^{*}$ & 0.081 (0.031)$^{*}$ & 0.045 (0.021)$^{*}$ \\
 & Rich AI & 0.0781 & 0.079 (0.027)$^{*}$ & 0.016 (0.032) & 0.063 (0.034) \\
 & Survey AI & 0.0798 & 0.103 (0.025)$^{*}$ & 0.039 (0.038) & 0.064 (0.035) \\
\addlinespace[2pt]
Foreign language & CFPS & 0.0768 & 0.079 (0.030)$^{*}$ & 0.067 (0.036) & 0.012 (0.013) \\
 & Rich AI & 0.0768 & 0.050 (0.024)$^{*}$ & 0.034 (0.028) & 0.017 (0.016) \\
 & Survey AI & 0.0773 & 0.055 (0.020)$^{*}$ & 0.022 (0.027) & 0.033 (0.016)$^{*}$ \\
\addlinespace[2pt]
Weekly hours & CFPS & 0.0746 & $-$0.240 (0.038)$^{*}$ & $-$0.105 (0.037)$^{*}$ & $-$0.135 (0.016)$^{*}$ \\
 & Rich AI & 0.0582 & $-$0.075 (0.036)$^{*}$ & $-$0.009 (0.031) & $-$0.066 (0.018)$^{*}$ \\
 & Survey AI & 0.0637 & $-$0.145 (0.034)$^{*}$ & $-$0.049 (0.031) & $-$0.096 (0.020)$^{*}$ \\
\addlinespace[2pt]
Management & CFPS & 0.0803 & 0.113 (0.025)$^{*}$ & 0.056 (0.025)$^{*}$ & 0.057 (0.012)$^{*}$ \\
 & Rich AI & 0.0775 & 0.051 (0.012)$^{*}$ & 0.026 (0.026) & 0.026 (0.025) \\
 & Survey AI & 0.0786 & 0.065 (0.015)$^{*}$ & 0.010 (0.025) & 0.056 (0.020)$^{*}$ \\
\bottomrule\end{tabular}}
\par\medskip\noindent\textit{Sample}\par\smallskip
\fittab{\begin{tabular}{lcc}\toprule
Construct & N & J \\ \midrule
Computer use & 6413 & 304 \\
Foreign language & 6413 & 304 \\
Weekly hours & 6774 & 306 \\
Management & 6409 & 304 \\
\bottomrule\end{tabular}}
\par\medskip\noindent\textit{Incremental $R^2$ over controls}\par\smallskip
\fittab{\begin{tabular}{lccc}\toprule
Construct & CFPS & Rich AI & Survey AI \\ \midrule
Computer use & 0.0040 & 0.0022 & 0.0039 \\
Foreign language & 0.0010 & 0.0009 & 0.0014 \\
Weekly hours & 0.0205 & 0.0040 & 0.0096 \\
Management & 0.0047 & 0.0018 & 0.0029 \\
\bottomrule\end{tabular}}
\par\smallskip\begin{minipage}{0.95\textwidth}\footnotesize Incremental $R^2$ is the $R^2$ of the model containing the individual measure and its occupation mean minus the controls-only $R^2$ of Table~\ref{tab:t3}, computed from unrounded values. Coefficients are standardized; $\beta_G=\beta_C+\beta_I$. Controls entered once. Occupation means are computed on each model's estimation sample. Standard errors in parentheses, clustered by occupation. Coefficients marked $^{*}$ are significant at the 5\% level (two-sided).\end{minipage}
\end{table}

\subsection{Joint contextual validation under simultaneous missingness}
The preceding validation exercises consider one survey dimension at a time. In those settings, a missing construct can be inferred using other observed respondent and job-characteristic information. Many practical applications, however, involve multiple missing dimensions simultaneously. In such settings, computer use cannot be used to predict management responsibilities, management responsibilities cannot be used to predict weekly hours, and so forth, because all of these dimensions are themselves unavailable.

To evaluate this more demanding scenario, we estimate a joint contextual model in which all four validated dimensions (computer use, foreign-language use, weekly hours, and management responsibilities) are treated as simultaneously unobserved. We impose a joint exclusion rule: any variable excluded when validating one dimension is excluded from the requests used to generate all four AI measures, and each dimension is still scored in its own request. In the present application, this removes direct or closely related measures of computer use, foreign-language use, weekly hours, management responsibilities, supervisory status, and number of subordinates from all four AI scores. After applying the rule, the information common to all four requests consists primarily of occupation, education degree, employer type, age, organizational size, and urban residence, with some additional dimension-specific information where it is not part of the union of exclusions. This design more closely resembles a realistic survey setting in which several job-characteristic measures were never collected. It is conceptually related to recent work evaluating LLM-based survey imputation under different missingness mechanisms \citep{holtdirk2026}; our design addresses a different validation target, namely whether the resulting AI-derived measures recover contextual coefficient structure rather than individual item responses alone.

Table~\ref{tab:t5} reports the results. The AI measures raise $R^2$ less than the CFPS measures, the survey-prompt AI more than the rich-prompt AI (row $\Delta R^2$). Weekly hours keeps a significant negative within-occupation association for all three measures, and a significant negative between-occupation association for CFPS and the survey-prompt AI but not for the rich prompt. Management keeps a significant positive group association for all three measures and a significant positive individual association for CFPS and the survey-prompt AI. Computer use keeps a significant group association for CFPS and the survey-prompt AI. No contextual coefficient is significant for CFPS or for either AI measure. The coefficient vectors and fitted values of the AI and CFPS models are closely related, more so under the survey prompt (Table~\ref{tab:t6}). Overall, the AI measures recover the broad directions of the contextual structure, mostly with attenuation, and the approach becomes less reliable when several related concepts are simultaneously unavailable. It performs best when a missing concept can be inferred from other observed respondent characteristics, and should be viewed primarily as a tool for recovering a limited number of missing constructs, not as an exact replacement for a jointly missing battery of related survey questions.

\begin{table}[htbp]\centering\footnotesize\renewcommand{\thetable}{5}
\caption{Joint contextual validation under simultaneous missingness (AI4)}\label{tab:t5}
\par\medskip\noindent\textit{Coefficients}\par\smallskip
\fittab{\begin{tabular}{llccc}\toprule
Dimension & Method & $\beta_G$ (SE) & $\beta_C$ (SE) & $\beta_I$ (SE) \\ \midrule
Computer use & CFPS & 0.090 (0.034)$^{*}$ & 0.052 (0.037) & 0.038 (0.020) \\
 & Rich AI4 & 0.079 (0.043) & 0.080 (0.052) & $-$0.002 (0.038) \\
 & Survey AI4 & 0.079 (0.033)$^{*}$ & 0.022 (0.050) & 0.057 (0.040) \\
\addlinespace[2pt]
Foreign language & CFPS & 0.046 (0.028) & 0.035 (0.032) & 0.011 (0.013) \\
 & Rich AI4 & $-$0.020 (0.041) & $-$0.027 (0.045) & 0.007 (0.017) \\
 & Survey AI4 & 0.001 (0.022) & $-$0.016 (0.027) & 0.016 (0.015) \\
\addlinespace[2pt]
Weekly hours & CFPS & $-$0.146 (0.047)$^{*}$ & 0.002 (0.046) & $-$0.148 (0.014)$^{*}$ \\
 & Rich AI4 & $-$0.052 (0.038) & 0.042 (0.041) & $-$0.094 (0.014)$^{*}$ \\
 & Survey AI4 & $-$0.105 (0.039)$^{*}$ & 0.001 (0.040) & $-$0.106 (0.015)$^{*}$ \\
\addlinespace[2pt]
Management & CFPS & 0.072 (0.023)$^{*}$ & 0.018 (0.023) & 0.054 (0.011)$^{*}$ \\
 & Rich AI4 & 0.039 (0.015)$^{*}$ & 0.016 (0.027) & 0.024 (0.025) \\
 & Survey AI4 & 0.045 (0.014)$^{*}$ & $-$0.014 (0.027) & 0.059 (0.021)$^{*}$ \\
\bottomrule\end{tabular}}
\par\medskip\noindent\textit{Model fit}\par\smallskip
\fittab{\begin{tabular}{lccc}\toprule
Metric & CFPS & Rich AI4 & Survey AI4 \\ \midrule
Controls-only $R^2$ & 0.0538 & 0.0538 & 0.0538 \\
Full $R^2$ & 0.0830 & 0.0639 & 0.0693 \\
$\Delta R^2$ & 0.0292 & 0.0101 & 0.0155 \\
\bottomrule\end{tabular}}
\par\medskip\noindent\textit{Sample}\par\smallskip
\fittab{\begin{tabular}{lcc}\toprule
Method & N & J \\ \midrule
CFPS & 6517 & 307 \\
Rich AI4 & 6517 & 307 \\
Survey AI4 & 6517 & 307 \\
\bottomrule\end{tabular}}
\par\smallskip\begin{minipage}{0.95\textwidth}\footnotesize Coefficients are standardized; $\beta_G=\beta_C+\beta_I$; controls entered once; standard errors in parentheses, clustered by occupation. All four dimensions are treated as simultaneously unobserved and scored in separate requests. Coefficients marked $^{*}$ are significant at the 5\% level (two-sided).\end{minipage}
\end{table}

\begin{table}[htbp]\centering\footnotesize\renewcommand{\thetable}{6}
\caption{Similarity of joint contextual results under simultaneous missingness}\label{tab:t6}
\par\medskip\noindent\textit{AI versus CFPS}\par\smallskip
\fittab{\begin{tabular}{p{5.2cm}cccc}\toprule
Measure & \shortstack{Rich vs CFPS\\($\beta_G,\beta_I$)} & \shortstack{Rich vs CFPS\\($\beta_C,\beta_I$)} & \shortstack{Survey vs CFPS\\($\beta_G,\beta_I$)} & \shortstack{Survey vs CFPS\\($\beta_C,\beta_I$)} \\ \midrule
Correlation of eight focal coefficients & 0.890 & 0.782 & 0.967 & 0.892 \\
RMSE of eight focal coefficients & 0.050 & 0.038 & 0.029 & 0.029 \\
Mean absolute difference of eight focal coefficients & 0.042 & 0.033 & 0.025 & 0.023 \\
Correlation of focal predicted components & 0.623 & 0.623 & 0.649 & 0.649 \\
RMSE of focal predicted components (centered) & 0.158 & 0.158 & 0.161 & 0.161 \\
Correlation of full fitted satisfaction & 0.844 & 0.844 & 0.838 & 0.838 \\
RMSE of full fitted satisfaction & 0.155 & 0.155 & 0.159 & 0.159 \\
\bottomrule\end{tabular}}
\par\medskip\noindent\textit{Rich versus survey prompt}\par\smallskip
\fittab{\begin{tabular}{p{5.2cm}cc}\toprule
Measure & \shortstack{Rich vs Survey\\($\beta_G,\beta_I$)} & \shortstack{Rich vs Survey\\($\beta_C,\beta_I$)} \\ \midrule
Correlation of eight focal coefficients & 0.906 & 0.706 \\
RMSE of eight focal coefficients & 0.032 & 0.037 \\
Mean absolute difference of eight focal coefficients & 0.024 & 0.032 \\
Correlation of focal predicted components & 0.888 & 0.888 \\
RMSE of focal predicted components (centered) & 0.087 & 0.087 \\
Correlation of full fitted satisfaction & 0.962 & 0.962 \\
RMSE of full fitted satisfaction & 0.072 & 0.072 \\
\bottomrule\end{tabular}}
\par\smallskip\begin{minipage}{0.95\textwidth}\footnotesize Same final sample as Table~\ref{tab:t5} ($N=6{,}517$, $J=307$). Focal coefficient vectors contain eight entries in fixed dimension order; the parenthesized pair indicates which two coefficients per dimension are compared. Prediction-based statistics are identical across the two parameterizations. RMSE of components is centered. Fit statistics are in-sample.\end{minipage}
\end{table}

\subsection{Sign concordance, attenuation and detectability}
All quantities in this subsection and in Table~\ref{tab:t6a} are relative to the CFPS coefficient, which is itself an estimate based on an imperfect survey item; they describe agreement with the survey benchmark, not errors with respect to the concept's true association. Tables~\ref{tab:t4} and~\ref{tab:t5} together contain 48 comparisons between an AI coefficient and the corresponding CFPS coefficient: each of the 24 CFPS coefficients (four dimensions, three coefficients, two tables) is compared with the AI coefficient under each of the two prompt styles. Table~\ref{tab:t6a} cross-tabulates their significance states and summarizes them. In none of the 19 comparisons in which the AI coefficient was significant at the 5\% level did it have a sign opposite to a significant CFPS estimate, that is, the benchmark-relative analogue of a Type S (sign) error \citep{gelman2014} did not occur; in 18 of the 19 the CFPS coefficient was also significant, and in the remaining one (the individual coefficient of foreign language under the survey prompt in Table~\ref{tab:t4}) the CFPS estimate has the same sign. These 19 comparisons involve 12 distinct CFPS coefficients. This is a descriptive count from a small number of related comparisons, and we do not use it to estimate an error rate. Significant AI coefficients rarely exceeded the corresponding CFPS coefficient in size, so the benchmark-relative analogue of a Type M (magnitude) error was also rare. Of the 30 comparisons in which the CFPS coefficient was significant (15 significant CFPS coefficients, each compared under both prompt styles), the AI coefficient was also significant with the same sign in 18: 12 of 14 group associations, 6 of 10 individual associations and none of 6 contextual associations (Table~\ref{tab:t6a}, lower block). Where the signs of the two coefficients disagree, neither is significant; such cases occur only in the joint model and mostly concern foreign language and contextual terms. These comparisons are not independent, since the two prompt styles share a model and feature set, the dimensions share respondents, and the three coefficients of a dimension are linked by $\beta_G=\beta_C+\beta_I$. The sign criterion detects reversals of direction but not losses of significance or size: an AI coefficient that is not significant, or much smaller than a significant CFPS coefficient, does not count as a conflict, since attenuation and limited sample size can produce such differences even when the direction is right (Table~\ref{tab:t6a}). We also do not know how the criterion would behave in other datasets or for other constructs.

\begin{table}[htbp]\centering\footnotesize\renewcommand{\thetable}{6A}
\caption{Scorecard of sign agreement between AI and CFPS coefficients}\label{tab:t6a}
\par\medskip\noindent\textit{Cross-tabulation of significance states (48 comparisons, Tables~\ref{tab:t4} and~\ref{tab:t5})}\par\smallskip
\fittab{\begin{tabular}{lcccc}\toprule
 & \multicolumn{3}{c}{AI coefficient} & \\ \cmidrule(lr){2-4}
CFPS coefficient & Sig.\ positive & Sig.\ negative & Not significant & Total \\ \midrule
Significantly positive & 11 & 0 & 9 & 20 \\
Significantly negative & 0 & 7 & 3 & 10 \\
Not significant & 1 & 0 & 17 & 18 \\
\midrule Total & 12 & 7 & 29 & 48 \\
\bottomrule\end{tabular}}
\par\medskip\noindent\textit{Summary by subset}\par\smallskip
\fittab{\begin{tabular}{lccccc}\toprule
Subset & \shortstack{Comparisons with\\significant CFPS\\coefficient} & \shortstack{Sign-\\detection} & \shortstack{Sign-\\conflict} & \shortstack{Sign-\\surplus} & \shortstack{Size ratio\\(median $|\mathrm{AI}|/|\mathrm{CFPS}|$)} \\ \midrule
All comparisons & 30 & 18 (60\%) & 0/19 & 1/19 & 0.61 \\
\addlinespace[2pt]
Rich prompt & 15 & 7 (47\%) & 0/7 & 0/7 & 0.46 \\
Survey prompt & 15 & 11 (73\%) & 0/12 & 1/12 & 0.71 \\
\addlinespace[2pt]
Between-group ($\beta_G$) & 14 & 12 (86\%) & 0/12 & 0/12 & 0.63 \\
Within-group ($\beta_I$) & 10 & 6 (60\%) & 0/7 & 1/7 & 0.71 \\
Contextual ($\beta_C$) & 6 & 0 (0\%) & 0/0 & 0/0 & 0.33 \\
\addlinespace[2pt]
Single-dimension (Table~\ref{tab:t4}) & 20 & 11 (55\%) & 0/12 & 1/12 & 0.54 \\
Joint, AI4 (Table~\ref{tab:t5}) & 10 & 7 (70\%) & 0/7 & 0/7 & 0.67 \\
\bottomrule\end{tabular}}
\par\smallskip\begin{minipage}{0.95\textwidth}\footnotesize All quantities are relative to the CFPS coefficients, not to the concepts' true associations. 5\% level; significance states $+1$, $-1$, $0$. \emph{Sign-detection}: comparisons with a significant CFPS coefficient in which the AI coefficient is significant with the same sign. \emph{Sign-conflict}: significant AI coefficients whose CFPS coefficient is significant with the opposite sign, out of all significant AI coefficients. \emph{Sign-surplus}: significant AI coefficients whose CFPS coefficient is not significant. A surplus need not indicate an error, since the two results may reconcile when more data become available. \emph{Size ratio}: median of $|\hat b_{\mathrm{AI}}|/|\hat b_{\mathrm{CFPS}}|$ over comparisons with a significant CFPS coefficient. The quantities are analogues of, not equivalents to, those of \citet{gelman2014}. The comparisons are dependent (the two prompts share a model and features; the dimensions share respondents).\end{minipage}
\end{table}

The pattern is one of attenuation without reversal: the median ratio of the AI to the CFPS coefficient is 0.61 (Table~\ref{tab:t6a}, size ratio). Detectability of an individual association depends on the size of the association, the attenuation, and the standard error of the AI coefficient. No AI contextual coefficient ($\beta_C$) is significant in Table~\ref{tab:t4} or Table~\ref{tab:t5}. A coefficient that is significant for one measure and not for the other does not by itself show that the two coefficients differ.

\section{Application to Latent Dimensions}\label{sec:latent}
The latent-dimension application should be interpreted differently from the CFPS validation analysis. For autonomy, people--things, creative--routine, and technology, CFPS does not contain direct validation measures. The analysis therefore cannot compare the AI-derived variables with survey measures of these constructs. Instead, it asks whether two independently designed prompting strategies yield similar contextual conclusions and whether the resulting associations are substantively plausible. The validated CFPS dimensions provide the methodological check; the latent dimensions illustrate how the approach can be applied when direct validation is unavailable.

\subsection{Measures and the treatment of time flexibility}
All latent measures reported in the main text are generated without the CFPS time-flexibility variable (qg604); the feature sets are in Table~\ref{tab:t10}. This exclusion was decided after an initial analysis, so we describe it here. When autonomy was scored with qg604, the measure added almost no explanatory power for satisfaction, unlike the version without it (Table~\ref{tab:t7}). Descriptive features of CFPS suggest why qg604 may not measure employee autonomy. The original qg604 code decreases with time flexibility; mean job satisfaction was 3.714 for qg604 = 3, 3.807 for qg604 = 2 and 3.681 for qg604 = 1, so the category coded as having the greatest flexibility did not have the greatest mean satisfaction, and the overall correlation between time flexibility and satisfaction was approximately zero. In the most flexible category, 42.7\% of respondents were self-employed or worked in a family business, compared with 6.4\% among respondents in the other two categories; that group was also less likely to have an employment contract, cash or material benefits, or work-related insurance, more likely to work every weekend or be on call, and had lower mean annual work income. In CFPS, high time flexibility can therefore accompany self-employment, irregular scheduling, and limited employment protection, not conventional employee autonomy alone. Because the choice was made after inspecting results, the latent findings should be read as exploratory.

\subsection{Single-dimension results}
Table~\ref{tab:t7} gives the single-dimension results, with significant coefficients marked. Autonomy and people--things show positive group and individual associations under both prompt styles, with contextual coefficients close to zero and the largest increases in $R^2$ among the four dimensions. Creative--routine shows a positive group association under the rich prompt only, and technology shows no association. No contextual association is established. Since $\beta_C=\beta_G-\beta_I$, a contextual coefficient is small when the group and individual associations are similar, as they are for autonomy and people--things.

\begin{table}[htbp]\centering\footnotesize\renewcommand{\thetable}{7}
\caption{Single-dimension results for the latent dimensions}\label{tab:t7}
\par\medskip\noindent\textit{Final single-dimension results}\par\smallskip
\fittab{\begin{tabular}{lcccccc}\toprule
Dimension & Prompt & $\beta_G$ (SE) & $\beta_C$ (SE) & $\beta_I$ (SE) & Baseline $R^2$ & Full $R^2$ \\ \midrule
Autonomy & Rich & 0.100 (0.025)$^{*}$ & $-$0.003 (0.028) & 0.102 (0.021)$^{*}$ & 0.0740 & 0.0803 \\
 & Survey & 0.117 (0.020)$^{*}$ & 0.018 (0.029) & 0.099 (0.024)$^{*}$ & 0.0740 & 0.0810 \\
\addlinespace[2pt]
People--things & Rich & 0.097 (0.019)$^{*}$ & $-$0.007 (0.039) & 0.104 (0.038)$^{*}$ & 0.0740 & 0.0810 \\
 & Survey & 0.087 (0.019)$^{*}$ & $-$0.039 (0.042) & 0.126 (0.036)$^{*}$ & 0.0740 & 0.0807 \\
\addlinespace[2pt]
Creative--routine & Rich & 0.053 (0.019)$^{*}$ & 0.063 (0.040) & $-$0.010 (0.040) & 0.0740 & 0.0759 \\
 & Survey & 0.038 (0.024) & 0.030 (0.041) & 0.007 (0.043) & 0.0740 & 0.0750 \\
\addlinespace[2pt]
Technology & Rich & $-$0.010 (0.023) & 0.005 (0.031) & $-$0.016 (0.028) & 0.0740 & 0.0740 \\
 & Survey & 0.036 (0.027) & 0.020 (0.032) & 0.016 (0.028) & 0.0740 & 0.0744 \\
\bottomrule\end{tabular}}
\par\medskip\noindent\textit{Sample and incremental fit}\par\smallskip
\fittab{\begin{tabular}{lcccc}\toprule
Dimension & Prompt & N & J & $\Delta R^2$ \\ \midrule
Autonomy & Rich & 6896 & 309 & 0.0064 \\
 & Survey & 6896 & 309 & 0.0071 \\
\addlinespace[2pt]
People--things & Rich & 6896 & 309 & 0.0070 \\
 & Survey & 6896 & 309 & 0.0067 \\
\addlinespace[2pt]
Creative--routine & Rich & 6896 & 309 & 0.0019 \\
 & Survey & 6896 & 309 & 0.0010 \\
\addlinespace[2pt]
Technology & Rich & 6896 & 309 & 0.0001 \\
 & Survey & 6896 & 309 & 0.0004 \\
\bottomrule\end{tabular}}
\par\smallskip\begin{minipage}{0.95\textwidth}\footnotesize Autonomy and creative--routine are scored without the time-flexibility variable. Coefficients are standardized; standard errors in parentheses, clustered by occupation; controls entered once; hours of work is a control. Baseline $R^2$ is the controls-only model; $\Delta R^2$ is the increase when the dimension is added. Coefficients marked $^{*}$ are significant at the 5\% level (two-sided).\end{minipage}
\end{table}

\subsection{Joint four-dimension analysis}
Because the four dimensions describe related aspects of work, their single-dimension associations may reflect variation shared across dimensions. We therefore estimate joint models that include all four dimensions simultaneously (Table~\ref{tab:t8}). As before, $\beta_G$ is the coefficient of the occupation mean in the deviation parameterization, $\beta_C$ the coefficient of the occupation mean in the raw parameterization, and $\beta_I$ the coefficient of the respondent-level deviation, with $\beta_G=\beta_C+\beta_I$. Together the four dimensions raise $R^2$ more than any single dimension does.

\emph{People--things.} The group association remains positive and significant under both prompts, somewhat smaller than in the single-dimension models. The individual association shrinks and is no longer significant once the other dimensions are included, and neither are the contextual coefficients.

\emph{Autonomy.} The individual association is almost unchanged, and the group association remains positive and significant under both prompts, though less precisely estimated under the rich prompt. The contextual coefficients are not significant. Autonomy is therefore associated with job satisfaction at both levels, without evidence of an additional contextual association.

\emph{Creative--routine and technology.} Neither dimension has a significant association under either prompt style once the other dimensions are included (Table~\ref{tab:t8}). The only significant single-dimension association of the two, the group association of creative--routine under the rich prompt (Table~\ref{tab:t7}), may therefore reflect variation shared with autonomy and people--things.

\emph{Consistency across prompt styles.} Table~\ref{tab:t9} compares the joint results from the two prompt styles. The eight group and individual coefficients are highly correlated across prompts, the contextual and individual coefficients less so, and the full fitted satisfaction is almost identical. As in the validation exercise, agreement is stronger for the $(\beta_G,\beta_I)$ representation than for $(\beta_C,\beta_I)$, so the contextual decomposition should be interpreted with greater caution than the group decomposition. Prompt variation affects the allocation of the fitted association among dimensions and especially among contextual components more than it affects the overall fitted outcome.

\emph{Findings that remain consistent.} Two conclusions hold under both prompt styles and in the single- and joint-dimension analyses: people--things has a positive group association, and autonomy has positive group and individual associations. These are reasonable to understand: occupational differences in people orientation matter for job satisfaction, and within-occupation differences in autonomy also matter, given two jobs with the same occupational mean autonomy.

\emph{Differences from version 1.} Two joint-model conclusions differ from version 1. First, autonomy's group association is now positive and significant in the joint model, whereas version 1 found no independent group association once the other dimensions were included; its individual association is positive in both versions. Second, technology no longer shows the positive group association that version 1 reported in the joint model. The positive group association of people--things is present in both versions, though smaller here. Because the latent dimensions have no survey target, these differences cannot arise from leakage of target variables. They may reflect the change of scoring model, the change from joint to single-dimension requests (which also changes the information available to each concept and removes any interaction among concepts within a request), or the instability of joint group coefficients for correlated dimensions. We cannot separate these explanations with the present data. The joint-model differences for autonomy and technology should therefore be treated as exploratory.

\begin{table}[htbp]\centering\footnotesize\renewcommand{\thetable}{8}
\caption{Joint four-dimension analysis of the latent dimensions}\label{tab:t8}
\par\medskip\noindent\textit{Coefficients}\par\smallskip
\fittab{\begin{tabular}{lcccc}\toprule
Dimension & Prompt & $\beta_G$ (SE) & $\beta_C$ (SE) & $\beta_I$ (SE) \\ \midrule
Autonomy & Rich & 0.084 (0.041)$^{*}$ & $-$0.016 (0.042) & 0.100 (0.021)$^{*}$ \\
 & Survey & 0.145 (0.035)$^{*}$ & 0.054 (0.041) & 0.092 (0.025)$^{*}$ \\
\addlinespace[2pt]
People--things & Rich & 0.078 (0.019)$^{*}$ & 0.022 (0.039) & 0.055 (0.037) \\
 & Survey & 0.081 (0.016)$^{*}$ & 0.008 (0.042) & 0.074 (0.039) \\
\addlinespace[2pt]
Creative--routine & Rich & $-$0.003 (0.024) & 0.029 (0.041) & $-$0.032 (0.037) \\
 & Survey & $-$0.043 (0.029) & $-$0.027 (0.047) & $-$0.015 (0.042) \\
\addlinespace[2pt]
Technology & Rich & $-$0.028 (0.025) & $-$0.024 (0.036) & $-$0.003 (0.029) \\
 & Survey & $-$0.026 (0.030) & $-$0.038 (0.038) & 0.012 (0.028) \\
\bottomrule\end{tabular}}
\par\medskip\noindent\textit{Model fit}\par\smallskip
\fittab{\begin{tabular}{lcc}\toprule
Metric & Prompt & $R^2$ \\ \midrule
Controls-only $R^2$ & Rich & 0.0740 \\
Controls-only $R^2$ & Survey & 0.0740 \\
Full $R^2$ & Rich & 0.0852 \\
Full $R^2$ & Survey & 0.0863 \\
$\Delta R^2$ & Rich & 0.0112 \\
$\Delta R^2$ & Survey & 0.0123 \\
\bottomrule\end{tabular}}
\par\medskip\noindent\textit{Sample}\par\smallskip
\fittab{\begin{tabular}{lcc}\toprule
Prompt & N & J \\ \midrule
Rich & 6896 & 309 \\
Survey & 6896 & 309 \\
\bottomrule\end{tabular}}
\par\smallskip\begin{minipage}{0.95\textwidth}\footnotesize Coefficients are standardized; standard errors in parentheses, clustered by occupation; controls entered once; hours of work is a control. Coefficients marked $^{*}$ are significant at the 5\% level (two-sided).\end{minipage}
\end{table}

\begin{table}[htbp]\centering\footnotesize\renewcommand{\thetable}{9}
\caption{Similarity of joint latent results across prompt styles}\label{tab:t9}
\par\medskip\noindent\textit{Similarity}\par\smallskip
\fittab{\begin{tabular}{lcc}\toprule
Similarity measure & G,I & C,I \\ \midrule
Correlation of eight focal coefficients & 0.9145 & 0.6993 \\
RMSE of eight focal coefficients & 0.0280 & 0.0340 \\
Mean absolute difference of eight focal coefficients & 0.0204 & 0.0265 \\
Correlation of latent-component predictions & 0.8921 & 0.8921 \\
RMSE of latent-component predictions (centered) & 0.0632 & 0.0632 \\
Correlation of full fitted satisfaction & 0.9779 & 0.9779 \\
RMSE of full fitted satisfaction & 0.0616 & 0.0616 \\
\bottomrule\end{tabular}}
\par\smallskip\begin{minipage}{0.95\textwidth}\footnotesize Same sample as Table~\ref{tab:t8} ($N=6{,}896$, $J=309$). Coefficient vectors follow the order autonomy, people--things, creative--routine, technology, with two coefficients per dimension.\end{minipage}
\end{table}

\section{Discussion}
Our results support three conclusions about respondent-level AI measures for contextual analysis.

First, when rich respondent and job information is available, such measures recover occupation-level structure well and individual-level structure partially. In the joint contextual model the AI measures add less explanatory power than the survey measures. However, the AI measures still detect the within-occupation association of weekly hours under both prompt styles (Tables~\ref{tab:t4} and~\ref{tab:t5}), and the survey-prompt individual AI measure explains more than any occupation-level measure, including the survey occupation mean (Table~\ref{tab:t3}).

Second, in our validation the AI measures behaved as conservative detectors. Across the 48 comparisons in Tables~\ref{tab:t4} and~\ref{tab:t5}, none of the 19 AI coefficients that were significant had a sign opposite to the survey coefficient, and 18 of them were also significant in the survey results, so the associations the AI measures flagged were almost always also found with the survey measures. These 19 comparisons involve only 12 distinct survey coefficients, so this is a descriptive count and not an estimate of an error rate. The price is sensitivity: the median ratio of the AI to the survey coefficient is 0.61, and the AI measures detected 18 of the 30 comparisons with a significant survey coefficient, none of the six contextual ones (Table~\ref{tab:t6a}). Used as a single-dimension screening tool, and counting an association only when both prompt styles find it, the AI measures point to positive occupation-level associations of computer use, foreign-language use and management with job satisfaction, and to negative occupation-level and individual associations of weekly hours (Table~\ref{tab:t4}); the survey items confirm all of these, and show in addition positive individual associations of computer use and management and contextual associations of computer use, weekly hours and management. These comparisons are few and correlated, a nonsignificant survey coefficient does not establish a true null, and we do not know how far the pattern carries over to other datasets or constructs. The survey-style prompt performed better than the rich construct prompt in all four dimensions.

Third, usefulness depends on the information available to the model. When the information is restricted to occupation and basic demographic characteristics, validation performance deteriorates sharply, and the within-occupation component is the part that largely disappears. The method uses information already present in the dataset to infer omitted constructs; it does not create information from nothing. Performance also declines when several related dimensions are simultaneously unobserved and therefore cannot be used to predict one another.

The latent-dimension application illustrates the intended use and its limits. Across prompt versions and in both single- and joint-dimension analyses, people--things has a positive occupation-level association with job satisfaction, and autonomy has positive associations at both the occupation level and the individual (within-occupation) level, while technology and creative--routine show no stable association once the other dimensions are included.

\paragraph{Practical guidance.} The evidence supports using respondent-level AI measures as a screening tool, not as a substitute for direct measurement.
\begin{center}
\fbox{\parbox{0.93\textwidth}{\small
(1) Give most weight to single-dimension between-group associations, then within-group associations, that have the same sign and are clearly significant under both prompt styles. Treat contextual coefficients and joint-model group coefficients with caution. In our validation every association meeting this criterion was also found with the survey measures.\\[2pt]
(2) Read a non-significant result as inconclusive, not as evidence of no association.\\[2pt]
(3) Expect a significant AI coefficient to have the direction that the corresponding survey coefficient would have, but usually a smaller size.\\[2pt]
(4) Confirm important findings with a second model, request design or sample, and supply rich job-specific information to the model.\\[4pt]
Use these measures to decide what deserves direct survey measurement, not to settle whether an association exists or how large it is.}}
\end{center}

\paragraph{Limitations.} First, the AI scores depend on how they are generated: version 1 used GPT-4o-mini and this version a locally run Qwen model, we do not exhaust model, prompt or time variation, and even with the same model and prompt, small differences can arise from the inference configuration. Second, the evidence rests on one survey, one country and one occupational coding, and only weekly hours has a strong individual-level benchmark. Third, the survey benchmarks are limited. The survey items are themselves imperfect, so agreement with them is not agreement with ground truth. The latent dimensions have no survey benchmark at all, and their treatment of time flexibility was chosen after inspecting results. Fourth, the results refer to the standard specification with the controls entered once. Sensitivity analyses that split the controls into within-occupation deviations and occupation means, or that add job features used in the prompts to the controls, are left to future work. Fifth, the comparisons across coefficients are few and correlated. Finally, respondent-level AI measures are not substitutes for direct survey measurement; they are most defensible for recovering a limited number of theoretically important concepts from rich existing respondent information.

As language models become more capable, the agreement reported here may improve, within the limits of the information supplied to the model. The validation framework, comparing the coefficients of AI measures with those of survey items and summarizing them by sign-detection, sign-conflict, sign-surplus and size ratio, applies to any survey in which some items can serve as benchmarks, and can be used to monitor this progress across models, prompts and surveys.

\section*{Acknowledgment}
Wenxin Jiang is partly supported by a discretionary fund from Northwestern University. He thanks the School of Social and Behavioral Sciences of Nanjing University for hosting his summer visits. Yuxiao Wu was partly supported by the Major Project of the National Social Science Fund of China (grant number 22\&ZD188). The data are from the China Family Panel Studies (CFPS), funded by the 985 Program of Peking University and carried out by the Institute of Social Science Survey of Peking University.

\section*{Data-Use and Generative AI Disclosure}
The AI scores in version 1 were generated with GPT-4o-mini through an API, before the CFPS Policy on the Use of AI Tools for CFPS Data (2 September 2026). In accordance with that policy, the scores in this version were generated from the 2022 CFPS public-use data with a locally deployed open-weight model (Qwen3.6-35B-A3B) on a computer that was physically disconnected from the internet during scoring. All inference took place on that computer, with no remote API calls or cloud services, and neither CFPS microdata nor the derived scores were transmitted to any external AI platform. The absence of network connections was checked by the operator at the start and end of each scoring session and by automated system checks throughout (Appendix~\ref{app:E}).

Large language models were used as part of the research methodology to generate respondent-level contextual measures from survey and occupational information, through the predefined prompting and scoring procedure described in the paper. Xuyang Wang used an LLM during the research. For this revision, Xuyang Wang used ChatGPT (OpenAI), and no other generative AI tool, to assist with coding and with drafting, editing and reviewing the manuscript; no CFPS microdata or AI-generated scores were shared with it. Yuxiao Wu did not use AI tools for writing or coding for this paper. Wenxin Jiang used generative AI tools during the research and manuscript preparation process: Microsoft Copilot, and, for this revision, Claude (Anthropic). They were used for tasks including coding assistance, drafting and editing text, brainstorming ideas, and checking references. For this revision, Claude helped write analysis code that Wenxin Jiang ran in a secure university computing environment; only aggregated results (coefficients, standard errors, correlations and summary statistics) were shared with the tool, and no respondent-level CFPS data were provided to it. The authors reviewed AI-assisted output and take responsibility for its use in the paper.

\appendix
\section{Variables and Analytic Roles}\label{app:A}
This appendix documents the variables used throughout the empirical analyses. Let $Y_i$ denote job satisfaction, $g(i)$ occupation membership, $X_i$ the regression control vector, and $W_{ki}$, $k=1,2,3,4$, the four observed CFPS measures used for validation. The regression control vector $X_i$ is conceptually distinct from the AI feature sets $F_k$ described in Appendix~\ref{app:B}. The control vector is used only in the regressions, whereas the feature sets determine the information supplied to the language model when generating respondent-level AI measures. Some variables, such as age, gender, education and urban residence, appear in both. Throughout the reported analyses, the default control vector was
\[
\begin{aligned}
X=\{&\text{age, gender, education years, urban residence, agricultural hukou,}\\
&\text{married indicator, public/state-owned employment, outdoor work,}\\
&\text{self-reported health, log income}\}.
\end{aligned}
\]
For analyses involving weekly hours as the focal validated construct, weekly hours was excluded from the control vector to avoid controlling for the construct being evaluated. For other analyses (including the latent dimensions), weekly hours was included as an additional control because it was not itself a focal construct.

Table~\ref{tab:t11} (Appendix~\ref{app:E}) lists every variable with its role and its construction for regression and for AI measurement.

\paragraph{Occupation titles.} Occupation titles were supplied to the language model whenever available and served as the primary contextual signal in AI measurement. For all labelled CFPS variables, including occupation codes, we extracted the human-readable category labels from the metadata attached to the survey variables. Occupation codes (\texttt{qg303code}) were therefore represented in the prompts by their corresponding occupation titles rather than by numeric codes.

\section{AI Feature Sets and Prompt Protocols}\label{app:B}
\subsection{Target concepts}
The framework considers eight job-related concepts: $U_1$ = computer use, $U_2$ = foreign-language use, $U_3$ = weekly hours, $U_4$ = management responsibility, $U_5$ = autonomy, $U_6$ = people orientation, $U_7$ = creative work, $U_8$ = technology intensity. The first four have corresponding CFPS survey measures $W_1,\dots,W_4$ and are used for validation. The remaining four have no direct validation counterparts and are treated as latent dimensions.

\subsection{Prompt styles}
For each concept $U_k$, two AI measures were generated, $Z^{\mathrm{rich}}_{ki}$ and $Z^{\mathrm{survey}}_{ki}$. The rich-prompt version asked the model to evaluate a broader conceptual construct. For example, the rich computer-use prompt asked how much a person's job requires computers, digital tools, office software, information systems, data processing, programming, or computer-based communication. The survey-prompt version instead asked the model to predict how the respondent would answer a survey-style question. For example, the survey computer-use prompt asked whether the respondent uses a computer for the current job, while the survey foreign-language prompt asked whether the respondent uses a foreign language for the current job. Agreement between the rich and survey versions provides evidence regarding robustness of the resulting AI measures.

\subsection{Response scales}
Computer use, foreign-language use, management responsibility, autonomy, people orientation, creative work, and technology intensity were measured on five-point scales. Weekly hours was treated differently: rather than generating a five-point rating, the AI model was asked to estimate the respondent's average weekly work hours, so weekly hours entered subsequent analyses as a continuous variable.

\subsection{Feature sets and exclusion rules}
For each concept $k$, let $F^{\mathrm{base}}_k$ denote the concept-specific feature set and $E_k$ the corresponding exclusion set. The main specification uses $F^{\mathrm{AI1}}_k=F^{\mathrm{base}}_k\setminus E_k$. Exclusion rules are concept specific. In this version each concept is scored in its own request, so a variable excluded for concept $k$ is absent from every request for that concept, while remaining available for other concepts unless explicitly excluded there as well. In version 1 the concepts were scored in a single request, so a variable excluded for one concept could be visible in the profile blocks of other concepts.

For the four validated concepts ($k=1,\dots,4$), a simultaneous-missingness specification was also considered. Define $E^{\mathrm{AI4}}=E_1\cup E_2\cup E_3\cup E_4$. The AI4 feature sets are then $F^{\mathrm{AI4}}_k=F^{\mathrm{base}}_k\setminus E^{\mathrm{AI4}}$, $k=1,\dots,4$. Thus any variable excluded for computer use, foreign-language use, weekly hours, or management responsibility is excluded from all four AI4 prediction tasks. Finally, a reduced-information specification $F^{\mathrm{reduced}}$, consisting of occupation, age, gender, education degree, marital status, and urban residence, was used for sensitivity analysis. This specification removes essentially all job-specific respondent information beyond occupation membership and basic demographics.

\subsection{Concept-specific feature sets}
Table~\ref{tab:t10} summarizes the role of each variable in the concept-specific feature sets.

\begin{table}[htbp]\centering\footnotesize\renewcommand{\thetable}{10}
\caption{Concept-specific feature sets: features available to each concept's own request}\label{tab:t10}
\fittab{\begin{tabular}{lcccccccc}\toprule
Feature & Comp & Lang & Hours & Mgmt & Auto & People & Creative & Tech \\ \midrule
Occupation & F & F & F & F & F & F & F & F \\
Age & F & F & F & F & F & F & F & F \\
Gender & -- & -- & F & F & -- & -- & -- & -- \\
Education degree & F & F & F & F & F & F & F & F \\
Urban residence & F & F & F & F & -- & -- & -- & -- \\
Marital status & -- & -- & F & F & -- & -- & -- & -- \\
Employer type & F & F & F & F & F & F & F & F \\
Industry & -- & -- & F & F & -- & -- & -- & -- \\
Work location & F & -- & -- & -- & -- & -- & -- & F \\
Province & -- & F & -- & -- & -- & -- & -- & -- \\
Organization size & F & F & F & F & F & -- & F & F \\
Promotion & -- & -- & F & F & F & -- & F & -- \\
Expected promotion & -- & -- & F & F & -- & -- & -- & -- \\
Income & -- & -- & F & F & -- & -- & -- & -- \\
Labor contract & -- & -- & F & F & -- & -- & -- & -- \\
Job tenure & -- & -- & F & F & -- & -- & -- & -- \\
Party membership & -- & -- & -- & F & F & -- & -- & -- \\
Management position & F & F & F & E4 & F & F & F & F \\
Direct reports & F & -- & F & E4 & F & F & F & -- \\
Number of subordinates & F & -- & F & E4 & F & -- & -- & -- \\
Computer use & E1 & -- & -- & -- & -- & -- & -- & F \\
Foreign-language use & -- & E2 & -- & -- & -- & -- & -- & F \\
Weekly hours & -- & -- & E3 & -- & F & -- & -- & -- \\
Time flexibility (initial analysis only) & -- & -- & -- & -- & F & -- & F & -- \\
Night shift & -- & -- & F & -- & F & -- & -- & -- \\
Weekend work & -- & -- & F & -- & F & -- & F & -- \\
On-call work & -- & -- & F & -- & F & -- & F & -- \\
\bottomrule\end{tabular}}
\par\smallskip\begin{minipage}{0.95\textwidth}\footnotesize F: included as a feature in $F_k^{\mathrm{base}}$; E1--E4: excluded because of leakage protection ($E_k$, for computer use, foreign-language use, weekly hours and management responsibility); --: unused for that concept. Comp = computer use; Lang = foreign-language use; Mgmt = management responsibility; Auto = autonomy; People = people--things; Creative = creative--routine; Tech = technology intensity. In the simultaneous-missingness design the union $E^{\mathrm{AI4}}=E_1\cup E_2\cup E_3\cup E_4$ is removed from all four validation concepts. Each concept is scored in its own request, so a variable excluded for one concept is absent from every request for that concept. Time flexibility is used only in the initial autonomy and creative--routine analysis (Section~\ref{sec:latent}); the main analyses exclude it.\end{minipage}
\end{table}

\section{CFPS Validation Measures}\label{app:C}
The first four concepts, $U_1,\dots,U_4$, have corresponding observed CFPS validation measures $W_1,\dots,W_4$, generated from survey questions $q_1,\dots,q_4$ administered in the CFPS occupational module.

\paragraph{Computer use.} The validation measure $W_1$ is based on question QG19: ``Do you use a computer for your current job?'' Response categories: 1. Yes; 5. No (recoded to 1 = yes, 0 = no; Table~\ref{tab:t10}).

\paragraph{Foreign-language use.} The validation measure $W_2$ is based on question QG18: ``Do you use a foreign language for your current job?'' Response categories: 1. Yes; 5. No (recoded to 1 = yes, 0 = no; Table~\ref{tab:t10}).

\paragraph{Weekly hours.} The validation measure $W_3$ is based on question QG6: ``Excluding lunch break, but including paid or unpaid extra working hours, how many hours per week on average did you work for this job in the past 12 months?'' Responses are recorded as average weekly work hours; the questionnaire instructs interviewers to convert minutes to hours and retain one decimal place.

\paragraph{Management responsibility.} The validation measure $W_4$ is based on question QG14: ``Do you have management duty for this job?'' The CFPS interviewer instructions define management duty as an occupation in the organization that has an official management function, such as the chief of a section, the director of a department, the head of a bureau, or a manager. Response categories: 1. Yes; 5. No (recoded to 1 = yes, 0 = no; Table~\ref{tab:t10}).

\section{Prompt Definitions}\label{app:D}
For each concept $U_k$, two prompt styles were used, $p\in\{\text{rich},\text{survey}\}$. The rich version asks the language model to evaluate a broader conceptual construct; the survey version asks the model to predict how the respondent would answer a survey-style question.

\subsection{Rich prompt definitions}
\begin{description}[leftmargin=0pt,labelsep=0.5em,itemsep=2pt]
\item[Computer use ($U_1$).] How much does this person's job require computers, digital tools, office software, information systems, data processing, programming, or computer-based communication? Scale: 1 = not at all; 2 = a little; 3 = moderately; 4 = a lot; 5 = very much.
\item[Foreign-language use ($U_2$).] How much does this person's job require use of a foreign language for reading, writing, speaking, translation, interpretation, international communication, or foreign-language documents? Scale: 1 = not at all; 2 = a little; 3 = moderately; 4 = a lot; 5 = very much.
\item[Weekly hours ($U_3$).] Excluding lunch break, but including paid or unpaid extra working hours, how many hours per week on average did this person work for this job in the past 12 months? Consider regular work schedules, overtime work, workload, managerial responsibilities, supervisory duties, workload intensity, promotion pressure, schedule demands, and expected time commitment. Return the number of work hours per week.
\item[Management responsibility ($U_4$).] CFPS survey item: Does this person have management duty for this job? Management duty: occupation in the organization that has official management function such as the chief of a section, the director of a department, the head of a bureau, manager, and so on. Scale: 1 = definitely no; 2 = probably no; 3 = uncertain; 4 = probably yes; 5 = definitely yes.
\item[Autonomy ($U_5$).] How much freedom, discretion, independence, self-direction, and decision-making authority does this person likely have over how work is performed? Scale: 1 = no freedom; 2 = little freedom; 3 = moderate freedom; 4 = much freedom; 5 = very much freedom.
\item[People orientation ($U_6$).] Is this person's job more people-oriented or things-oriented? People-oriented work involves teaching, helping, caring, persuading, serving, coordinating, managing, or communicating with people. Things-oriented work involves machines, tools, equipment, materials, production processes, physical systems, or technical operations. Scale: 1 = entirely things-oriented; 2 = mostly things-oriented; 3 = mixed; 4 = mostly people-oriented; 5 = entirely people-oriented.
\item[Creative work ($U_7$).] Is this person's job more creative/nonroutine or routine/standardized? Creative work involves imagination, new ideas, flexible judgment, problem solving, artistic production, or innovation. Routine work involves standardized, repetitive, predictable tasks following fixed procedures. Scale: 1 = highly routine; 2 = mostly routine; 3 = mixed; 4 = mostly creative/nonroutine; 5 = highly creative/nonroutine.
\item[Technology intensity ($U_8$).] How technologically intensive is this person's job overall? Technology-intensive work involves digital technology, engineering, scientific equipment, automation, electronics, information systems, or technical expertise. Scale: 1 = not technological; 2 = slightly technological; 3 = moderately technological; 4 = highly technological; 5 = extremely technological.
\end{description}

\subsection{Survey prompt definitions}
Except for weekly hours, each survey prompt follows the question with ``Based on the available information, predict this person's likely answer.'' and the scale 1 = definitely no; 2 = probably no; 3 = uncertain; 4 = probably yes; 5 = definitely yes. The questions are:
\begin{description}[leftmargin=0pt,labelsep=0.5em,itemsep=2pt]
\item[$U_1$.] CFPS survey item: ``Do you use a computer for your current job?''
\item[$U_2$.] CFPS survey item: ``Do you use a foreign language for your current job?''
\item[$U_3$.] CFPS survey item: ``Excluding lunch break, but including paid or unpaid extra working hours, how many hours per week on average did you work for this job in the past 12 months?'' The request ended: ``Based on the available information, predict this person's most likely answer. Return the number of weekly work hours.''
\item[$U_4$.] Survey question: ``Do you have management duty for this job?''
\item[$U_5$.] Survey question: ``Does your job give you freedom in how work is done?''
\item[$U_6$.] Survey question: ``Does your work mainly involve people?''
\item[$U_7$.] Survey question: ``Does your work require creativity or new ideas?''
\item[$U_8$.] Survey question: ``Does your work require advanced technology?''
\end{description}

\section{Prompt Templates}\label{app:E}

All AI measures in this version were generated with Qwen3.6-35B-A3B. Each request scored one concept for one respondent under one prompt style (or, for occupation-title scores, one concept for one occupation title). Each request consisted of a single user message; no system prompt was used. The values supplied to the model for concept $k$ are the features in Table~\ref{tab:t10} (Appendix~\ref{app:B}). Below, braces mark placeholders; no template is shown with respondent data.

\subsection{Respondent-level message}
The user message was:
{\footnotesize
\begin{verbatim}
You are simulating responses to job-characteristic survey items, not job satisfaction.
Answer as a worker with the occupation and characteristics below.
Do not infer overall satisfaction, happiness, or mental health.

Prompt version: {style}
Survey/profile year: 2022

Return a value for the following job-characteristic item.
Use the scale specified in the item definition.

- {dim}: {definition}

Important instructions:
- Use the dimension-specific profile block below for this item.
- Occupation is always included when available.
- If occupation says "Occupation not reported", rely on the other respondent
  characteristics.
- Return ONLY valid JSON with exactly this key: {"{dim}": number}.
- Do not include explanations, markdown, or text outside the JSON object.

Dimension: {dim}
Pay particular attention to these variables for this dimension:
{profile}
\end{verbatim}}
\noindent Here \texttt{\{style\}} is \texttt{rich} or \texttt{survey}; \texttt{\{dim\}} is the concept key (\texttt{computer}, \texttt{foreign\_language}, \texttt{weekly\_hours}, \texttt{management}, \texttt{autonomy}, \texttt{people\_things}, \texttt{creative\_routine}, \texttt{technology}); \texttt{\{definition\}} is the item text of Appendix~\ref{app:D} for that style and concept; and \texttt{\{profile\}} is the concept-specific profile block. The line beginning ``If occupation says'' is a single line in the request.

\subsection{Profile blocks}
A profile block has one line per feature, of the form \texttt{Label: \{value\}}. The rich and survey styles used the same block for a given concept and design. Lines with unavailable values were omitted, except that the occupation line was kept with the value ``Occupation not reported''. For example, the block for computer use in the main design was:
{\footnotesize
\begin{verbatim}
Occupation title: {occupation_title}
Highest educational degree in CFPS2022: {education}
Work location: {work_location}
Type of employer: {employer_type}
Age: {age}
Number of employees that company/work unit/business has: {organization_size}
Do you have administrative/management position: {management_position}
Do you have any direct subordinates: {direct_reports}
Number of subordinates that you have: {number_of_subordinates}
Urban area (Census Bureau's definition): {urban}
\end{verbatim}}
\noindent The blocks for the other concepts have the same form and contain the features marked F in Table~\ref{tab:t10}. The designs differ only in the profile block; the rest of the message is the same.
\begin{itemize}[leftmargin=1.3em,itemsep=1pt,topsep=2pt]
\item \textit{Main design (AI1).} The block for concept $k$ contains $F^{\mathrm{AI1}}_k=F^{\mathrm{base}}_k\setminus E_k$. The latent measures of autonomy and creative--routine reported in the main text use blocks without time flexibility (Section~\ref{sec:latent}).
\item \textit{Simultaneous missingness (AI4).} For the four validated concepts, the block contains $F^{\mathrm{AI4}}_k=F^{\mathrm{base}}_k\setminus E^{\mathrm{AI4}}$, with $E^{\mathrm{AI4}}=E_1\cup E_2\cup E_3\cup E_4$; that is, the management position, direct reports and number of subordinates are removed from the blocks of computer use, foreign-language use and weekly hours.
\item \textit{Reduced information.} For the four validated concepts, the block contains only occupation, age, gender, education degree, marital status and urban residence.
\end{itemize}
In version 1, by contrast, a single request contained the profile blocks of all eight concepts and asked for values of all of them.

\subsection{Occupation-title messages}
Occupation-title scores were generated for the four validated concepts, with the occupation title as the only input. For computer use the message was:
{\footnotesize
\begin{verbatim}
Rate the Chinese occupation title on one job-characteristic scale.
Use only the typical work content of the occupation. Do not answer job satisfaction.

Occupation title: {occupation_title}
Question: How much does this occupation require computers, digital tools, office
software, information systems, data processing, programming, or computer-based
communication?
Scale: 1 = not at all; 2 = a little; 3 = moderately; 4 = a lot; 5 = very much.

Return ONLY valid JSON with exactly this key: {"computer": number}.
Decimals are allowed. Do not include explanations, markdown, or text outside the JSON
object.
\end{verbatim}}
\noindent The messages for foreign-language use and management responsibility have the same form, with the question and scale of the rich definition in Appendix~\ref{app:D} (with ``this occupation'' in place of ``this person's job'' in the foreign-language question, as in the computer-use message above). The question and the scale are each a single line in the request. For weekly hours the message was:
{\footnotesize
\begin{verbatim}
Estimate the typical weekly work hours of a Chinese occupation.

Use only the typical characteristics of the occupation.
Do not answer job satisfaction.

Occupation title: {occupation_title}

Question:
Excluding lunch breaks, but including paid and unpaid overtime,
how many hours per week does a typical worker in this occupation work?

Consider:
- regular work schedules
- overtime
- managerial responsibilities
- workload
- common industry practices

Return ONLY valid JSON with exactly this key: {"weekly_hours": number}.
Decimals are allowed. Do not include explanations, markdown, or text outside the JSON
object.
\end{verbatim}}

\subsection{Output format, decoding and computing environment}
A strict JSON schema, enforced during decoding, required a single numeric value keyed by the requested concept. Decoding was greedy (temperature 0, \texttt{top\_p} $=1$, \texttt{top\_k} $=0$, \texttt{min\_p} $=0$, no presence, frequency or repetition penalty; seed 20261017), with at most 1{,}024 generated tokens per request. Thinking mode and streaming were disabled; the Qwen chat template wrapped the user message and added the assistant prefix with an empty thinking block. Requests were processed serially, one request in flight with batch size 1 for both prefill and decoding. Multi-token prediction, speculative decoding, Apple Neural Engine offload, key--value cache quantization and reuse of prompt prefixes across requests were disabled. The same checkpoint and settings were used for the respondent-level and occupation-title requests.

The checkpoint was \texttt{Jundot/\allowbreak Qwen3.6-35B-A3B-oQ8e-mtp}, a quantization of Qwen3.6-35B-A3B with predominantly 8-bit affine weights (8.59 effective bits per weight on average), at Hugging Face revision\\[2pt] \centerline{\texttt{911d4eac3d62cfb964f8dd0e0b0096fd343983f5}.}\\[-2pt] \noindent The hashes of all weight files matched the run manifest. Inference used MLX-VLM 0.7.10 and MLX 0.32.4 with Python 3.12.9 on a 16-inch MacBook Pro with an Apple M4 Max chip and 64 GB of unified memory, under macOS 27.2.1. The model was loaded from local storage, and all requests were handled on the same computer. Before and after each of the two scoring sessions, the operator confirmed that the only attached cable was the power cable and that Wi-Fi, Bluetooth and sharing services were disabled. Throughout each session, automated checks every 10 seconds recorded the network interfaces, routes and sockets and the attached devices; none of these checks failed.

\subsection{Common processing and cleaning rules}
Several coded categorical variables were converted into human-readable text before prompt construction, including schedule flexibility, night-shift frequency, and weekend-work frequency. Weekly hours was treated as a quantitative construct and generated directly as a numerical estimate; all other concepts were measured on five-point scales. Labelled variables were first converted to ordinary numeric values for regression. CFPS special missing-value codes were treated as missing. Binary yes/no variables were converted to 1--0 indicators. Nonnegative quantitative variables were restricted to nonnegative values. Invalid occupation codes were removed before constructing occupation-based variables and matching occupation titles.

Table~\ref{tab:t11} lists the variables used in the analysis and in AI measurement.

{\renewcommand{\thetable}{11}\footnotesize
\begin{longtable}{>{\raggedright\arraybackslash}p{3.0cm}>{\raggedright\arraybackslash}p{1.4cm}>{\raggedright\arraybackslash}p{4.8cm}>{\raggedright\arraybackslash}p{5.2cm}}
\caption{Variables used in the analysis and in AI measurement}\label{tab:t11}\\ \toprule
Variable & Role & Regression construction & AI representation \\ \midrule \endfirsthead
\multicolumn{4}{l}{\textit{Table 11 (continued)}}\\ \toprule Variable & Role & Regression construction & AI representation \\ \midrule \endhead
\bottomrule \endfoot
Job satisfaction \texttt{qg406} & $Y$ & Five-point satisfaction scale; valid responses retained. & Not used. \\
Occupation \texttt{qg303code} & $g$, $F$ & Invalid occupation codes removed. & Occupation titles obtained from the CFPS value labels. \\
Computer use \texttt{qg19} & $W_1$, $F$ & Binary indicator (1 = yes, 0 = no). & Human-readable yes/no description. \\
Foreign-language use \texttt{qg18} & $W_2$, $F$ & Binary indicator (1 = yes, 0 = no). & Human-readable yes/no description. \\
Weekly hours \texttt{qg6} & $W_3$, $F$ & Finite nonnegative weekly hours. & Weekly hours information supplied directly. \\
Management duty \texttt{qg14} & $W_4$, $F$ & Binary indicator (1 = yes, 0 = no). & Human-readable yes/no description. \\
Age \texttt{age} & $X$, $F$ & Continuous age. & Age value supplied directly. \\
Gender \texttt{gender} & $X$, $F$ & Binary indicator. & Human-readable gender description. \\
Education degree \texttt{cfps2022edu} & $F$ & Not used. & Educational-attainment description. \\
Education years \texttt{cfps2022eduy} & $X$ & Finite nonnegative years. & Not used. \\
Marital status \texttt{qea0} & $X$, $F$ & Married indicator (1 = married, 0 = other status). & Marital-status description. \\
Hukou \texttt{qa301} & $X$ & Agricultural hukou indicator (1 = agricultural hukou, 0 = non-agricultural/residence hukou). & Not used. \\
Urban residence \texttt{urban22} & $X$, $F$ & Binary urban indicator. & Urban/rural description. \\
Self-rated health \texttt{qp201} & $X$ & Five-point scale reversed so larger values indicate better health. & Not used. \\
Income \texttt{incomeb} & $X$, $F$ & For nonnegative income, $\log(1+\text{income})$ is used in regressions. & Nonnegative income value is supplied to the language model. \\
Employer type \texttt{qg2} & $X$, $F$ & Public/SOE indicator: 1 = government, public institution, or state-owned enterprise. & Employer-type description. \\
Work location \texttt{qg20} & $X$, $F$ & Outdoor-work indicator. Transportation-based workplaces treated as missing. & Work-location description. \\
Organization size \texttt{qg16} & $F$ & Finite nonnegative value. & Organization-size information. \\
Promotion \texttt{qg15} & $F$ & Management, technical, or joint promotion coded as promotion. & Promotion history description. \\
Expected promotion \texttt{qg1501} & $F$ & Retained as categorical information. & Expected-promotion description. \\
Job tenure \texttt{qg2032} & $F$ & Binary indicator. & Job-tenure description. \\
Labor contract \texttt{qg5} & $F$ & Binary indicator. & Labor-contract description. \\
Direct reports \texttt{qg17} & $F$ & Binary indicator. & Direct-reports description. \\
Number of subordinates \texttt{qg1701} & $F$ & Finite nonnegative value. & Number-of-subordinates information. \\
Party membership \texttt{qn4001} & $F$ & Binary indicator. & Party-membership description. \\
Time flexibility \texttt{qg604} & $F$ & Not used in regressions. & Three schedule-flexibility categories converted into human-readable descriptions. Used only in the initial analysis. \\
Night shift \texttt{qg601} & $F$ & Not used in regressions. & Human-readable frequency description. \\
Weekend work \texttt{qg602} & $F$ & Not used in regressions. & Human-readable frequency description. \\
On-call work \texttt{qg603} & $F$ & Not used in regressions. & Human-readable description. \\
Industry \texttt{qg302code} & $F$ & Not used in regressions. & Industry description. \\
Work-location province (derived from \texttt{provcd22}, \texttt{qg301}, \texttt{qg301a\_code}, \texttt{jobclass}) & $F$ & Not used in regressions. & Work-location province in text. Uses the residential province (\texttt{provcd22}) when it is valid and either (1) \texttt{qg301} indicates that the work location is within mainland China and no valid alternative province (\texttt{qg301a\_code}) differs from \texttt{provcd22}, or (2) \texttt{jobclass} indicates family agricultural work. If a valid alternative work province (\texttt{qg301a\_code}) differs from \texttt{provcd22}, \texttt{qg301a\_code} takes precedence. \\
\end{longtable}}

\section{Occupation-Level Comparison Methods}\label{app:F}
To benchmark the respondent-level AI measures, we also considered two occupation-level approaches based solely on occupation titles. Unlike the respondent-level measures, these benchmarks do not use respondent-specific information and therefore assign the same score to all individuals within a given occupation. The list of occupation titles comes from the labels attached to the \texttt{qg303code} variable in the CFPS data.

\paragraph{Occupation-level AI scores.} The first benchmark uses direct AI scoring of occupation titles, with Qwen3.6-35B-A3B in this version (1,218 valid requests). For each occupation, the model evaluated the typical characteristics of that occupation using only the title as input, focusing on the occupation itself and not on job satisfaction. For the three ordinal concepts the model returned a score on a five-point scale using the same rich concept definitions employed in the respondent-level measures; for weekly hours it estimated the average weekly hours of a typical worker in the occupation. The scores were merged to individuals through occupation codes.

\paragraph{Occupation embedding scores.} The second benchmark uses occupation embeddings from the model text-embedding-3-large (unchanged from version 1). Each occupation title was converted into a vector representation, with the model asked to generate a neutral representation of the occupation's typical work content, responsibilities, and work setting. For each target concept, a semantic direction was constructed from positive and negative textual descriptions of opposite ends of the dimension (for example, computer-intensive versus non-computer work), and occupation scores were obtained by projecting each embedding onto that direction. Where occupation codes were duplicated in the source file (codes 60809, 61109 and 61809), the first source row was retained, as in version 1.

\end{document}